%% file: main.tex
\documentclass[10pt,twocolumn,letterpaper]{article}

\usepackage[pagenumbers]{cvpr} %

\usepackage{graphicx}
\usepackage{amsmath}
\usepackage{amssymb}
\usepackage{booktabs}
\usepackage{makecell}
\usepackage[export]{adjustbox}
\usepackage[normalem]{ulem}
\usepackage{textpos}  %

\usepackage{tabulary,multirow,overpic,xcolor,subfloat}
\usepackage[pagebackref=false, breaklinks=true, letterpaper=true, colorlinks,
            citecolor=citecolor, linkcolor=linkcolor, bookmarks=false]{hyperref}
\definecolor{citecolor}{HTML}{0071BC}
\definecolor{linkcolor}{HTML}{ED1C24}

\newcommand{\app}{\raise.17ex\hbox{$\scriptstyle\sim$}}

\def\x{\times}
\newcolumntype{x}[1]{>{\centering\arraybackslash}p{#1pt}}
\newcolumntype{y}[1]{>{\raggedright\arraybackslash}p{#1pt}}
\newcommand{\dt}[1]{\fontsize{5pt}{0.1em}\selectfont (#1)}
\newlength\savewidth\newcommand\shline{\noalign{\global\savewidth\arrayrulewidth
  \global\arrayrulewidth 1pt}\hline\noalign{\global\arrayrulewidth\savewidth}}
\newcommand{\tablestyle}[2]{\setlength{\tabcolsep}{#1}\renewcommand{\arraystretch}{#2}\centering\footnotesize}
\makeatletter\renewcommand\paragraph{\@startsection{paragraph}{4}{\z@}
  {.5em \@plus1ex \@minus.2ex}{-.5em}{\normalfont\normalsize\bfseries}}\makeatother

\DeclareMathAlphabet\mathbfcal{OMS}{cmsy}{b}{n}

\definecolor{Gray}{gray}{0.5}
\newcommand{\demph}[1]{\textcolor{Gray}{#1}}

\newcommand{\modelname}{Mask2Former\xspace}

\newcommand{\modelnamelong}{Masked-attention Mask Transformer\xspace}
\newcommand{\sota}[0]{state-of-the-art\xspace}

\usepackage[capitalize]{cleveref}
\crefname{section}{Sec.}{Secs.}
\Crefname{section}{Section}{Sections}
\Crefname{table}{Table}{Tables}
\crefname{table}{Tab.}{Tabs.}

\newcommand{\figref}[1]{Figure~\ref{#1}}
\newcommand{\secref}[1]{Section~\ref{#1}}
\newcommand{\tabref}[1]{Table~\ref{#1}}
\newcommand{\appref}[1]{Appendix~\ref{#1}}

\def\cvprPaperID{*****}
\def\confName{CVPR}
\def\confYear{2022}

\crefname{section}{\S}{\S\S}
\crefname{subsection}{\S}{\S\S}
\crefformat{table}{Table~#2#1#3}
\crefformat{figure}{Figure~#2#1#3}
\crefformat{equation}{Equation~(#2#1#3)}
\crefformat{algorithm}{Algorithm~#2#1#3}
\crefformat{appendix}{Appendix~#2#1#3}

\newcommand{\authorskip}{\hspace{2.5mm}}

\begin{document}

\title{\Large \modelnamelong for Universal Image Segmentation}

\author{
 Bowen Cheng$^{1,2}$\thanks{Work done during an internship at Facebook AI Research.} \authorskip Ishan Misra$^1$ \authorskip Alexander G. Schwing$^2$ \authorskip
 Alexander Kirillov$^1$ \authorskip Rohit Girdhar$^1$ \\
 $^1$Facebook AI Research (FAIR) \authorskip $^2$University of Illinois at Urbana-Champaign (UIUC) \\
{\small \url{https://bowenc0221.github.io/mask2former}}
}

\maketitle

\input{sections/abs}

\input{sections/intro}

\input{sections/relwork}

\input{sections/approach}
\input{sections/expts}

\input{sections/concl}

{
\footnotesize
{\bf \noindent Ethical considerations:} 
While our technical innovations do not appear to have any inherent biases, the models trained with our approach on real-world datasets should undergo ethical review to ensure the predictions do not propagate problematic stereotypes, and the approach is not used for applications including but not limited to illegal surveillance.

\noindent\textbf{Acknowledgments:} Thanks to Nicolas Carion and Xingyi Zhou for helpful feedback. BC and AS are supported in part by NSF \#1718221, 2008387, 2045586, 2106825, MRI \#1725729, NIFA 2020-67021-32799 and Cisco Systems Inc.\ (CG 1377144 - thanks for access to Arcetri).
}

{\small
\bibliographystyle{ieee_fullname}
\bibliography{refs}
}

\clearpage
\appendix
\begin{center}{\bf \Large Appendix}\end{center}\vspace{-2mm}
\renewcommand{\thetable}{\Roman{table}}
\renewcommand{\thefigure}{\Roman{figure}}
\setcounter{table}{0}
\setcounter{figure}{0}
\input{sections/appendix}

\end{document}

%% file: sections/abs.tex
\begin{abstract}

Image segmentation groups %
pixels with different semantics, \eg, category or instance membership. %
Each choice of semantics defines a task. %
While only the semantics of each task differ,
current research focuses on designing specialized architectures for each task.
We present \modelnamelong (\modelname), a new architecture capable of addressing any image segmentation task (panoptic, instance or semantic).
Its key components include \emph{masked attention}, which extracts localized features by constraining cross-attention within predicted mask regions. %
In addition to reducing the research effort by at least three times, it outperforms the best specialized architectures by a significant margin on four popular datasets.
Most notably, \modelname sets a new state-of-the-art for panoptic segmentation (57.8 PQ on COCO), instance segmentation (50.1 AP on COCO) and semantic segmentation (57.7 mIoU on ADE20K).
\end{abstract}

%% file: sections/intro.tex
\vspace{-2mm}
\section{Introduction}

Image segmentation studies the problem of grouping  pixels.
Different semantics for grouping pixels, \eg, category or instance membership, have led to different types of segmentation tasks, such as panoptic, instance or semantic segmentation.
While these tasks differ only in semantics, current methods develop specialized architectures for each task.
Per-pixel classification architectures
based on Fully Convolutional Networks (FCNs)~\cite{long2015fully} are used for semantic segmentation, while mask classification architectures~\cite{he2017mask,detr} that predict a set of binary masks each associated with a single category,  dominate instance-level segmentation.
Although such \emph{specialized} architectures~\cite{long2015fully,deeplabV3plus,he2017mask,chen2019hybrid} have advanced each individual task, they lack the flexibility to generalize to the other tasks. For example, FCN-based architectures struggle at instance segmentation, leading to the evolution of different architectures for instance segmentation compared to semantic segmentation.
Thus, duplicate research and (hardware) optimization effort is spent on each specialized architecture for every task.

\begin{figure}[t!]
    \centering
    \includegraphics[width=\linewidth]{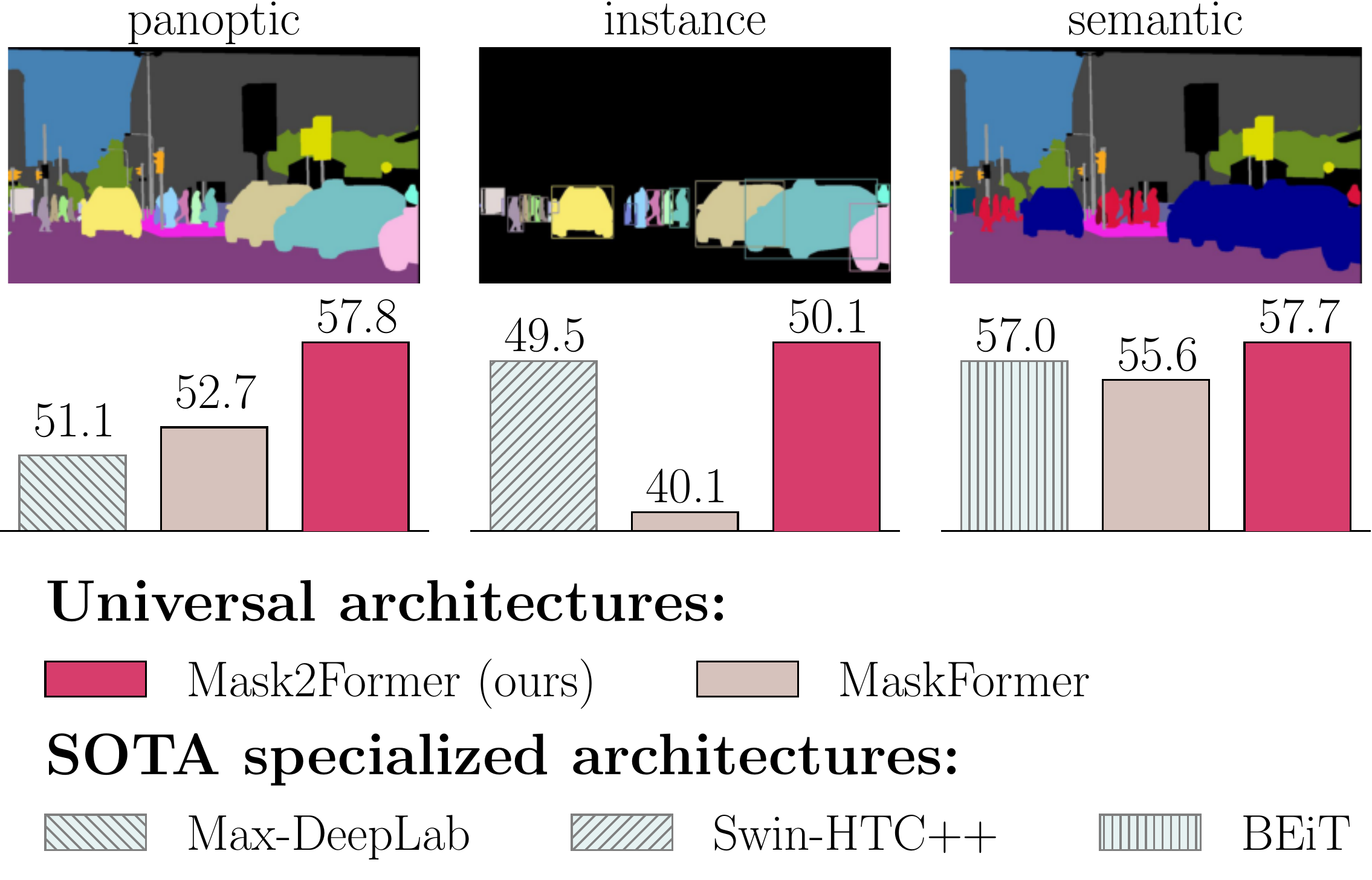}
    \caption{
        State-of-the-art segmentation architectures are typically specialized for each image segmentation task.
        Although recent work %
        has proposed universal architectures that attempt all tasks and are competitive on semantic and panoptic segmentation, they struggle with segmenting instances. We propose \textbf{\modelname}, which, for the first time, outperforms the best specialized architectures on three studied segmentation tasks on multiple datasets.
        }
    \label{fig:teaser}
    \vspace{-1mm}
\end{figure}

To address this fragmentation, %
recent work~\cite{cheng2021maskformer,zhang2021knet} has attempted to design \emph{universal architectures}, that are capable of addressing all segmentation tasks with the same architecture (\ie, universal image segmentation).
These architectures are typically based on an end-to-end set prediction objective (\eg, DETR~\cite{detr}), and successfully tackle multiple tasks without modifying the architecture, loss, or the training procedure. 
Note, universal architectures are still trained separately for different tasks and datasets, albeit having the same architecture. %
In addition to being flexible, universal architectures have recently shown \sota
results on semantic and panoptic segmentation~\cite{cheng2021maskformer}.
However, recent work  still focuses on advancing specialized architectures~\cite{fapn,strudel2021segmenter,QueryInst}, which raises the question: why haven't universal architectures replaced specialized ones?

Although existing universal architectures are flexible enough to tackle any segmentation task, %
as shown in \figref{fig:teaser}, in practice their performance %
lags behind the best specialized architectures.
For instance, the best reported performance of universal architectures~\cite{cheng2021maskformer,zhang2021knet}, is currently lower ($>9$ AP) than the SOTA specialized architecture for instance segmentation~\cite{chen2019hybrid}.
Beyond the inferior performance, universal architectures are also harder to train. They typically require more advanced hardware and a much longer training schedule. For example, training MaskFormer~\cite{cheng2021maskformer} takes 300 epochs to reach 40.1 AP and it can only fit a single image in a GPU with 32G memory. In contrast, the specialized Swin-HTC++~\cite{chen2019hybrid} obtains better performance in only 72 epochs. Both the performance and training efficiency issues hamper the deployment of universal architectures.

In this work, we propose a universal image segmentation architecture named \modelnamelong (\textbf{\modelname}) that outperforms specialized architectures across different segmentation tasks, while still being easy to train on every task.
We build upon a simple meta architecture~\cite{cheng2021maskformer} consisting of a
backbone feature extractor~\cite{he2016deep,liu2021swin}, a pixel decoder~\cite{lin2016feature} and a Transformer decoder~\cite{vaswani2017attention}.
We propose key improvements that enable better results and efficient training.
First, we use \emph{masked attention} in the Transformer decoder which restricts the attention to localized features centered around predicted segments, which can be either objects or regions depending on the specific semantic for grouping.
Compared to the  cross-attention used in a standard Transformer decoder which attends to all locations in an image, our masked attention leads to faster convergence and improved performance.
Second, we use \emph{multi-scale high-resolution features} which help the model to segment small objects/regions.
Third, we propose \emph{optimization improvements} such as switching the order of self and cross-attention, making query features learnable, and removing dropout; all of which improve performance without additional compute.
Finally, we save $3\x$ training memory without affecting the performance by \emph{calculating mask loss on few randomly sampled points}.
These improvements not only boost the model performance, but also make training significantly easier, making universal architectures more accessible to users with limited compute.

We evaluate \modelname on three image segmentation tasks (panoptic, instance and semantic segmentation) using four popular datasets (COCO~\cite{lin2014coco}, Cityscapes~\cite{Cordts2016Cityscapes}, ADE20K~\cite{zhou2017ade20k} and Mapillary Vistas~\cite{neuhold2017mapillary}).
For the first time, on all these benchmarks, our single architecture performs on par or better than specialized architectures.
\modelname sets the new state-of-the-art of \textbf{57.8 PQ} on COCO panoptic segmentation~\cite{kirillov2017panoptic}, \textbf{50.1 AP} on COCO instance segmentation~\cite{lin2014coco} and \textbf{57.7 mIoU} on ADE20K semantic segmentation~\cite{zhou2017ade20k} using the exact same architecture.

%% file: sections/relwork.tex
\section{Related Work}

\noindent\textbf{Specialized semantic segmentation architectures} typically treat the task as a per-pixel classification problem. FCN-based architectures~\cite{long2015fully} independently predict a category label for every pixel. Follow-up methods find context to play an important role for  precise per-pixel classification and focus on designing customized context modules~\cite{deeplabV2,deeplabV3,zhao2017pspnet} or self-attention variants~\cite{wang2018non,fu2019dual,yuan2018ocnet,huang2019ccnet,zheng2021rethinking,strudel2021segmenter}.

\noindent\textbf{Specialized instance segmentation architectures} are typically based upon ``mask classification.'' They predict a set of binary masks each associated with a single class label. The pioneering work, Mask R-CNN~\cite{he2017mask}, generates masks from detected bounding boxes. Follow-up methods either focus on detecting more precise bounding boxes~\cite{cai2018cascade,chen2019hybrid}, or finding new ways to generate a dynamic number of masks, \eg, using dynamic kernels~\cite{tian2020conditional,wang2020solov2,yolact-plus-arxiv2019} or clustering algorithms~\cite{kirillov2016instancecut,cheng2020panoptic}. Although the performance has been advanced in each task, these specialized innovations lack the flexibility to generalize from one to the other, leading to duplicated research effort. For instance, although multiple approaches have been proposed for building feature pyramid representations~\cite{lin2016feature}, as we show in our experiments,
BiFPN~\cite{tan2020efficientdet} performs better for instance segmentation while FaPN~\cite{fapn} performs better for semantic segmentation.

\noindent\textbf{Panoptic segmentation} has been proposed to unify both semantic and instance segmentation tasks~\cite{kirillov2017panoptic}. Architectures for panoptic segmentation either combine the best of specialized semantic and instance segmentation architectures into a single framework~\cite{xiong19upsnet,kirillov2019panopticfpn,cheng2020panoptic,li2021fully} or design novel objectives that equally treat semantic regions and instance objects~\cite{detr,wang2021max}. Despite those new architectures, researchers continue to develop specialized architectures for different image segmentation tasks~\cite{strudel2021segmenter,QueryInst}.
We find panoptic architectures usually only report performance on a single panoptic segmentation task~\cite{wang2021max}, which does not guarantee good performance on other tasks (\figref{fig:teaser}). For example, panoptic segmentation does not measure architectures' abilities to rank predictions as instance segmentations. Thus, we refrain from referring to architectures that are only evaluated for panoptic segmentation as universal architectures. Instead, here, we evaluate our \modelname on all studied tasks to guarantee generalizability.

\noindent\textbf{Universal architectures} have emerged with DETR~\cite{detr} and show that mask classification architectures with an end-to-end set prediction objective are general enough for any image segmentation task. 
MaskFormer~\cite{cheng2021maskformer} shows that mask classification based on DETR not only performs well on panoptic segmentation but also achieves state-of-the-art on semantic segmentation. K-Net~\cite{zhang2021knet} further extends set prediction to instance segmentation. Unfortunately, these architectures fail to replace specialized models as their performance on particular tasks or datasets is still worse than the best specialized architecture (\eg, MaskFormer~\cite{cheng2021maskformer} cannot segment instances well). To our knowledge, \modelname is the first
architecture that outperforms state-of-the-art specialized architectures on all considered tasks and datasets. %

%% file: sections/approach.tex
\section{\modelnamelong}
We now present \modelname.
We first review a  meta architecture for mask classification that \modelname is built upon.
Then, we introduce our new Transformer decoder with \emph{masked attention} which is the key to better convergence and results.
Lastly, we propose training improvements that make \modelname  efficient and accessible.

\subsection{Mask classification preliminaries}
Mask classification architectures group pixels into $N$ segments by predicting $N$ binary masks, along with $N$ corresponding category labels. Mask classification is sufficiently general to address any segmentation task by assigning different semantics, \eg, categories or instances, to different segments. However, the challenge is to find good representations for each segment. For example, Mask R-CNN~\cite{he2017mask} uses bounding boxes as the representation which limits its application to semantic segmentation.
Inspired by DETR~\cite{detr}, each segment in an image can be represented as a $C$-dimensional feature vector (``object query'') and can be processed by a Transformer decoder, trained with a set prediction objective.
A simple meta architecture would consist of three components.
A \emph{backbone} that extracts low-resolution features from an image. A \emph{pixel decoder} that gradually upsamples low-resolution features from the output of the backbone to generate high-resolution per-pixel embeddings. And finally a \emph{Transformer decoder} that operates on image features to process object queries. The final binary mask predictions are decoded from per-pixel embeddings with object queries. One successful instantiation of such a meta architecture is MaskFormer~\cite{cheng2021maskformer}, and we refer readers to \cite{cheng2021maskformer} for more details.

\subsection{Transformer decoder with masked attention}
\label{sec:method:arch:transformer_decoder}

\begin{figure}[t]
    \centering
    \includegraphics[width=\linewidth]{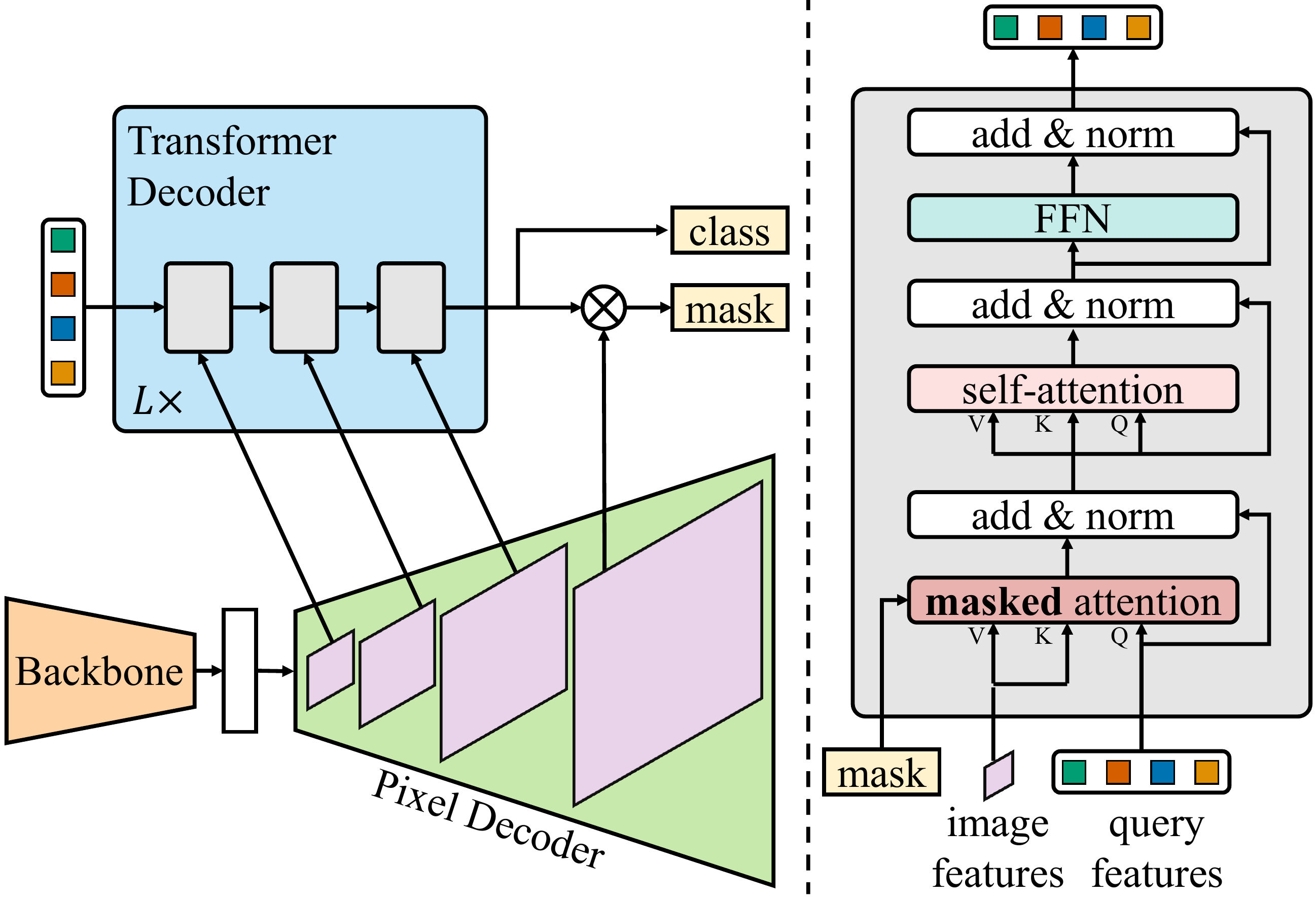}
    \caption{\textbf{\modelname overview.} \modelname adopts the same meta architecture as MaskFormer~\cite{cheng2021maskformer} with a backbone, a pixel decoder and a Transformer decoder. We propose a new Transformer decoder with \emph{masked attention} instead of the standard cross-attention (\secref{sec:method:arch:transformer_decoder:mask_attn}). To deal with small objects, we propose an efficient way of utilizing high-resolution features from a pixel decoder by feeding one scale of the multi-scale feature to one Transformer decoder layer at a time (\secref{sec:method:arch:transformer_decoder:ms_feat}). In addition, we switch the order of self and cross-attention (\ie, our masked attention), make query features learnable, and remove dropout to make computation more effective (\secref{sec:method:arch:transformer_decoder:other}). Note that positional embeddings and predictions from intermediate Transformer decoder layers are omitted in this figure for readability.}
    \label{fig:arch}
\end{figure}

\modelname adopts the aforementioned meta architecture, with our proposed Transformer decoder (\figref{fig:arch} right) replacing the standard one. %
The key components of our Transformer decoder include a \emph{masked attention} operator, which extracts  localized features by constraining cross-attention to within the foreground region of the predicted mask for each query, instead of
attending to the full feature map.
To handle small objects, we propose an efficient multi-scale strategy to utilize high-resolution features.
It feeds successive feature maps from the pixel decoder's feature pyramid %
into successive Transformer decoder layers in a round robin fashion.
Finally, we incorporate optimization improvements that boost model performance without introducing additional computation.
We now discuss these improvements in detail.

\subsubsection{Masked attention}
\label{sec:method:arch:transformer_decoder:mask_attn}

Context features have been shown to be important for image segmentation~\cite{deeplabV2,deeplabV3,zhao2017pspnet}. However, recent studies~\cite{sun2021rethinking,gao2021smca} suggest that the slow convergence of Transformer-based models is due to global context in the cross-attention layer, as it takes many training epochs for cross-attention to learn to attend to localized object regions~\cite{sun2021rethinking}. We hypothesize that local features are enough to update query features and context information can be gathered through self-attention. For this we propose \emph{masked attention}, a variant of cross-attention that only attends within the foreground region of the predicted mask for each query.

Standard cross-attention (with residual path) computes
\begin{align}
\mathbf{X}_{l} = \text{softmax}(\mathbf{Q}_{l}\mathbf{K}^{\text{T}}_{l})\mathbf{V}_{l} + \mathbf{X}_{l-1}.
\end{align}
Here, $l$ %
is the layer index, $\mathbf{X}_{l} \in \mathbb{R}^{N \x C}$ refers to $N$ $C$-dimensional query features at the $l^{\text{th}}$ layer and $\mathbf{Q}_{l} = f_Q(\mathbf{X}_{l-1}) \in \mathbb{R}^{N \x C}$. $\mathbf{X}_{0}$ denotes input query features to the Transformer decoder.
$\mathbf{K}_{l}, \mathbf{V}_{l} \in \mathbb{R}^{H_{l}W_{l} \x C}$
are the image features under transformation $f_K(\cdot)$ and $f_V(\cdot)$ respectively, and $H_l$ and $W_l$ are the spatial resolution of image features that we will introduce next in \secref{sec:method:arch:transformer_decoder:ms_feat}. $f_Q$, $f_K$ and $f_V$ are linear transformations.

Our masked attention modulates the attention matrix via
\begin{align}
  \mathbf{X}_{l} = \text{softmax}(\mathbfcal{M}_{l-1} + \mathbf{Q}_{l}\mathbf{K}^{\text{T}}_{l})\mathbf{V}_{l} + \mathbf{X}_{l-1}.
\end{align}
Moreover, the attention mask $\mathbfcal{M}_{l-1}$ at feature location $(x, y)$ is
\begin{align}
\mathbfcal{M}_{l-1}(x, y) = \left\{\begin{array}{ll}
  0  & \text{if~} \mathbf{M}_{l-1}(x,y)=1 \\
    -\infty & \text{otherwise}
\end{array}\right..
\end{align}
Here, $\mathbf{M}_{l-1} \in \{0,1\}^{N \x H_{l}W_{l}}$ is the binarized output (thresholded at $0.5$) of the resized mask prediction of the previous ($l-1$)-th Transformer decoder layer. It is resized to the same resolution of $\mathbf{K}_{l}$. $\mathbf{M}_{0}$ is the binary mask prediction obtained from  $\mathbf{X}_{0}$, \ie, before feeding query features into the Transformer decoder.

\vspace{-2mm}
\subsubsection{High-resolution features}
\label{sec:method:arch:transformer_decoder:ms_feat}

High-resolution features improve model performance, especially for small objects~\cite{detr}. However, this is computationally demanding. Thus, we propose an efficient multi-scale strategy to introduce high-resolution features while controlling the increase in computation. Instead of always using the high-resolution feature map, we utilize a feature pyramid which consists of both low- and high-resolution features and feed one resolution of the multi-scale feature to one Transformer decoder layer at a time.

Specifically, we use the feature pyramid produced by the \emph{pixel decoder} with resolution $1/32$, $1/16$ and $1/8$ of the original image. For each resolution, we add both a sinusoidal positional embedding $e_{\text{pos}} \in \mathbb{R}^{H_{l}W_{l} \x C}$, following~\cite{detr}, and a learnable scale-level embedding $e_{\text{lvl}} \in \mathbb{R}^{1 \x C}$, following~\cite{zhu2021deformable}. We use those, from lowest-resolution to highest-resolution for the corresponding Transformer decoder layer as shown in \figref{fig:arch} left. We repeat this 3-layer Transformer decoder  $L$ times.  Our final Transformer decoder hence has $3L$ layers. More specifically, the first three layers receive a feature map of resolution $H_{1} = H/32$, $H_{2} = H/16$, $H_{3} = H/8$ and $W_{1} = W/32$, $W_{2} = W/16$, $W_{3} = W/8$, where $H$ and $W$ are the original image resolution. This pattern is repeated in a round robin fashion for all following layers.

\subsubsection{Optimization improvements}
\label{sec:method:arch:transformer_decoder:other}
A standard Transformer decoder layer~\cite{vaswani2017attention} consists of three modules to process query features in the following order: a self-attention module, a cross-attention and a feed-forward network (FFN). Moreover, query features ($\mathbf{X}_{0}$) are \emph{zero initialized} before being fed into the Transformer decoder and are associated with \emph{learnable} positional embeddings. Furthermore, dropout is applied to both residual connections and attention maps.

To optimize the Transformer decoder design, we make the following three improvements. First, we switch the order of self- and cross-attention (our new ``masked attention'') to make computation more effective: query features to the first self-attention layer are image-independent and do not have signals from the image, thus applying self-attention is unlikely to enrich information. Second, we make query features ($\mathbf{X}_{0}$) learnable as well (we still keep the learnable query positional embeddings), and learnable query features are directly supervised before being used in the Transformer decoder to predict masks ($\mathbf{M}_{0}$). We find these learnable query features function like a region proposal network~\cite{Ren2015a} and have the ability to generate mask proposals. Finally, we find dropout is not necessary and usually decreases performance. We thus  completely remove dropout in our decoder.

\subsection{Improving training efficiency}
\label{sec:method:arch:loss}

One limitation of training universal architectures is the large memory consumption due to high-resolution mask prediction, making them less accessible than the more memory-friendly specialized architectures~\cite{he2017mask,chen2019hybrid}. For example, MaskFormer~\cite{cheng2021maskformer} can only fit a single image in a GPU with 32G memory. Motivated by PointRend~\cite{kirillov2020pointrend} and Implicit PointRend~\cite{cheng2021pointly}, which show a segmentation model can be trained with its mask loss calculated on $K$ randomly sampled points instead of the whole mask, we calculate the mask loss with sampled points in both the matching and the final loss calculation. More specifically, in the \emph{matching loss} that constructs the cost matrix for bipartite matching, we \emph{uniformly} sample the same set of $K$ points for all prediction and ground truth masks. In the \emph{final loss} between predictions and their matched ground truths, we sample different sets of $K$ points for different pairs of prediction and ground truth using \emph{importance sampling}~\cite{kirillov2020pointrend}. We set $K=12544$, \ie, $112 \x 112$ points. This new training strategy effectively reduces training memory by $3\x$, from 18GB to 6GB per image, making \modelname more accessible to users with limited computational resources.

%% file: sections/expts.tex
\section{Experiments}
We demonstrate \modelname is an effective architecture for  universal image segmentation  through comparisons with specialized state-of-the-art architectures on  standard benchmarks.
We evaluate %
our proposed design decisions through ablations on all three tasks.
Finally we show \modelname generalizes beyond the standard benchmarks,
obtaining state-of-the-art results on four datasets.

\begin{table*}[t]
  \centering

  \tablestyle{4pt}{1.2}\scriptsize\begin{tabular}{l | lcc | x{20}x{20}x{20} | x{20}x{24} |x{24}x{24}x{18}}
  method & backbone & query type & epochs & PQ & PQ$^\text{Th}$ & PQ$^\text{St}$ & AP$^\text{Th}_\text{pan}$ & mIoU$_\text{pan}$ & \#params. & FLOPs & fps \\
  \shline
  \demph{DETR~\cite{detr}} & \demph{R50} & \demph{100 queries} & \demph{500+25} & \demph{43.4} & \demph{48.2} & \demph{36.3} & \demph{31.1} & \demph{-} & \demph{-} & \demph{-} & \demph{-} \\
  MaskFormer~\cite{cheng2021maskformer} & R50 & 100 queries & 300 & 46.5 & 51.0 & 39.8 & 33.0 & 57.8 & \phantom{0}45M & \phantom{0}181G & 17.6 \\
  \textbf{\modelname} (ours) & R50 & 100 queries & 50 & \textbf{51.9} & \textbf{57.7} & \textbf{43.0} & \textbf{41.7} & \textbf{61.7} & \phantom{0}44M & \phantom{0}226G & \phantom{0}8.6 \\
  \hline
  \demph{DETR~\cite{detr}} & \demph{R101} & \demph{100 queries} & \demph{500+25} & \demph{45.1} & \demph{50.5} & \demph{37.0} & \demph{33.0} & \demph{-} & \demph{-} & \demph{-} & \demph{-} \\
  MaskFormer~\cite{cheng2021maskformer} & R101 & 100 queries & 300 & 47.6 & 52.5 & 40.3 & 34.1 & 59.3 & \phantom{0}64M & \phantom{0}248G & 14.0 \\
  \textbf{\modelname} (ours) & R101 & 100 queries & 50 & \textbf{52.6} & \textbf{58.5} & \textbf{43.7} & \textbf{42.6} & \textbf{62.4} & \phantom{0}63M & \phantom{0}293G & \phantom{0}7.2 \\
  \hline
  Max-DeepLab~\cite{wang2021max} & Max-L & 128 queries & 216 & 51.1 & 57.0 & 42.2 & - & - & 451M & 3692G & \demph{-} \\
  MaskFormer~\cite{cheng2021maskformer} & Swin-L$^{\text{\textdagger}}$ & 100 queries & 300 & 52.7 & 58.5 & 44.0 & 40.1 & 64.8 & 212M & \phantom{0}792G & \phantom{0}5.2 \\
  K-Net~\cite{zhang2021knet} & Swin-L$^{\text{\textdagger}}$ & 100 queries & 36 & 54.6 & 60.2 & 46.0 & - & - & - & - & - \\
  \textbf{\modelname} (ours) & Swin-L$^{\text{\textdagger}}$ & \emph{200 queries} & \emph{100} & \textbf{57.8} & \textbf{64.2} & \textbf{48.1} & \textbf{48.6} & \textbf{67.4} & 216M & \phantom{0}868G & \phantom{0}4.0 \\
  \end{tabular}
  \vspace{-1mm}

   \caption{\textbf{Panoptic segmentation on COCO panoptic \texttt{val2017} with 133 categories.} \modelname consistently outperforms MaskFormer~\cite{cheng2021maskformer} by a large margin with different backbones on all metrics. Our best model outperforms prior state-of-the-art MaskFormer by $5.1$ PQ and K-Net~\cite{zhang2021knet} by $3.2$ PQ.
Backbones pre-trained on ImageNet-22K are marked with $^{\text{\textdagger}}$.
}
\vspace{-3mm}

\label{tab:panseg:coco}
\end{table*}

\noindent\textbf{Datasets.}
We study \modelname %
using four widely used image segmentation datasets that support semantic, instance and panoptic segmentation:
COCO~\cite{lin2014coco} (80 ``things'' and 53 ``stuff'' categories), ADE20K~\cite{zhou2017ade20k} (100 ``things'' and 50 ``stuff'' categories), Cityscapes~\cite{Cordts2016Cityscapes} (8 ``things'' and 11 ``stuff'' categories) and Mapillary Vistas~\cite{neuhold2017mapillary} (37 ``things'' and 28 ``stuff'' categories). Panoptic and semantic segmentation tasks are evaluated on the union of ``things'' and ``stuff'' categories while instance segmentation is only evaluated on the ``things'' categories.

\noindent\textbf{Evaluation metrics.}
For \emph{panoptic segmentation}, we use the standard \textbf{PQ} (panoptic quality) metric~\cite{kirillov2017panoptic}. We further report \textbf{AP$^\text{Th}_\text{pan}$}, which is the AP evaluated on the ``thing'' categories using instance segmentation annotations, and \textbf{mIoU$_\text{pan}$}, which is the mIoU for semantic segmentation by merging instance masks from the same category, of the same model trained \emph{only} with panoptic segmentation annotations. For \emph{instance segmentation}, we use the standard \textbf{AP} (average precision) metric~\cite{lin2014coco}. For \emph{semantic segmentation}, we use \textbf{mIoU} (mean Intersection-over-Union)~\cite{everingham2015pascal}.

\begin{table*}[t]
  \centering

  \tablestyle{4pt}{1.2}\scriptsize\begin{tabular}{l| lcc | x{20}x{20}x{20}x{20} | x{30} |x{24}x{24}x{18}}
  method & backbone & query type & epochs & AP & AP$^\text{S}$ & AP$^\text{M}$ & AP$^\text{L}$ & AP$^\text{boundary}$ & \#params. & FLOPs & fps \\
  \shline
  MaskFormer~\cite{cheng2021maskformer} & R50 & 100 queries & 300 & 34.0 & 16.4 & 37.8 & 54.2 & 23.0 & \phantom{0}45M & \phantom{0}181G & 19.2 \\
  \demph{Mask R-CNN~\cite{he2017mask}} & \demph{R50} & \demph{dense anchors} & \demph{36} & \demph{37.2} & \demph{18.6} & \demph{39.5} & \demph{53.3} & \demph{23.1} & \demph{\phantom{0}44M} & \demph{\phantom{0}201G} & \demph{15.2} \\
  Mask R-CNN~\cite{he2017mask,ghiasi2021simple,du2021simple} & R50 & dense anchors & 400 & 42.5 & \textbf{23.8} & 45.0 & 60.0 & 28.0 & \phantom{0}46M & \phantom{0}358G & 10.3 \\
  \textbf{\modelname} (ours) & R50 & 100 queries & 50 & \textbf{43.7} & 23.4 & \textbf{47.2} & \textbf{64.8} & \textbf{30.6} & \phantom{0}44M & \phantom{0}226G & \phantom{0}9.7 \\
  \hline
  \demph{Mask R-CNN~\cite{he2017mask}} & \demph{R101} & \demph{dense anchors} & \demph{36} & \demph{38.6} & \demph{19.5} & \demph{41.3} & \demph{55.3} & \demph{24.5} & \demph{\phantom{0}63M} & \demph{\phantom{0}266G} & \demph{10.8} \\
  Mask R-CNN~\cite{he2017mask,ghiasi2021simple,du2021simple} & R101 & dense anchors & 400 & 43.7 & \textbf{24.6} & 46.4 & 61.8 & 29.1 & \phantom{0}65M & \phantom{0}423G & \phantom{0}8.6 \\
  \textbf{\modelname} (ours) & R101 & 100 queries & 50 & \textbf{44.2} & 23.8 & \textbf{47.7} & \textbf{66.7} & \textbf{31.1} & \phantom{0}63M & \phantom{0}293G & \phantom{0}7.8 \\
  \hline
  QueryInst~\cite{QueryInst} & Swin-L$^{\text{\textdagger}}$ & 300 queries & 50 & 48.9 & 30.8 & 52.6 & 68.3 & 33.5 & - & - & \phantom{0}\demph{3.3} \\
  Swin-HTC++~\cite{liu2021swin,chen2019hybrid} & Swin-L$^{\text{\textdagger}}$ & dense anchors & 72 & 49.5 & \textbf{31.0} & 52.4 & 67.2 & 34.1 & 284M & 1470G & - \\
  \textbf{\modelname} (ours) & Swin-L$^{\text{\textdagger}}$ & \emph{200 queries} & \emph{100} & \textbf{50.1} & 29.9 & \textbf{53.9} & \textbf{72.1} & \textbf{36.2} & 216M & \phantom{0}868G & \phantom{0}4.0 \\
  \end{tabular}
  \vspace{-1mm}

   \caption{\textbf{Instance segmentation on COCO \texttt{val2017} with 80 categories.} \modelname outperforms strong Mask R-CNN~\cite{he2017mask} baselines for both AP and AP$^\text{boundary}$~\cite{cheng2021boundary} metrics when training with $8\x$ fewer epochs. Our best model is also competitive to the state-of-the-art specialized instance segmentation model on COCO and has higher boundary quality. For a fair comparison, we only consider single-scale inference and models trained using only COCO \texttt{train2017} set data.
Backbones pre-trained on ImageNet-22K are marked with $^{\text{\textdagger}}$.}

\vspace{-3mm}

\label{tab:insseg:coco}
\end{table*}

\subsection{Implementation details}
\label{sec:exp:impl}
\noindent We adopt settings from~\cite{cheng2021maskformer} with the following differences:

\noindent\textbf{Pixel decoder.} \modelname is compatible with %
any existing pixel decoder module. %
In MaskFormer~\cite{cheng2021maskformer}, FPN~\cite{lin2016feature} is chosen as the default %
for its simplicity. Since our goal is to demonstrate strong performance across different segmentation tasks, %
we use the more advanced multi-scale deformable attention Transformer (MSDeformAttn)~\cite{zhu2021deformable} as our default pixel decoder. Specifically, we use 6 MSDeformAttn layers applied to feature maps with resolution $1/8$, $1/16$ and $1/32$,
and use a simple upsampling layer with lateral connection on the final %
$1/8$ feature map to generate the  feature map of resolution $1/4$ as the per-pixel embedding.
In our ablation study, we show that this pixel decoder provides best results across different segmentation tasks.

\noindent\textbf{Transformer decoder.} We use our Transformer decoder proposed in \secref{sec:method:arch:transformer_decoder} with $L=3$ (\ie, 9 layers total) and 100 queries by default. An auxiliary loss is added to every intermediate Transformer decoder layer and to the learnable query features before the Transformer decoder.

\noindent\textbf{Loss weights.} We use the binary cross-entropy loss (instead of focal loss~\cite{lin2017focal} in~\cite{cheng2021maskformer}) and the dice loss~\cite{milletari2016v} for our mask loss: $\mathcal{L}_{\text{mask}} = \lambda_{\text{ce}}\mathcal{L}_{\text{ce}} +\lambda_{\text{dice}}\mathcal{L}_{\text{dice}}$. We set $\lambda_{\text{ce}} = 5.0$ and $\lambda_{\text{dice}} = 5.0$. The final loss is a combination of mask loss and classification loss: $\mathcal{L}_{\text{mask}} + \lambda_{\text{cls}}\mathcal{L}_{\text{cls}}$ and we set $\lambda_{\text{cls}} = 2.0$ for predictions matched with a ground truth and $0.1$ for the ``no object,'' \ie, predictions that have not been matched with any ground truth.

\noindent\textbf{Post-processing.} We use the exact same post-processing as~\cite{cheng2021maskformer} to acquire the expected output format for panoptic and semantic segmentation from pairs of binary masks and class predictions. Instance segmentation requires additional confidence scores for each prediction. We multiply class confidence and mask confidence (\ie, averaged foreground per-pixel binary mask probability) for a final confidence. %

\subsection{Training settings}
\noindent\textbf{Panoptic and instance segmentation.} We use Detectron2~\cite{wu2019detectron2} and follow the updated Mask R-CNN~\cite{he2017mask} baseline settings\footnote{\scriptsize \url{https://github.com/facebookresearch/detectron2/blob/main/MODEL\_ZOO.md\#new-baselines-using-large-scale-jitter-and-longer-training-schedule}} for the COCO dataset. More specifically, we use AdamW~\cite{loshchilov2018decoupled} optimizer and the step learning rate  schedule. We use an initial learning rate of $0.0001$ and a weight decay of $0.05$ for all backbones. A learning rate multiplier of $0.1$ is applied to the backbone and we decay the learning rate at 0.9 and 0.95 fractions of the total number of training steps by a factor of 10. If not stated otherwise, we train our models for 50 epochs with a batch size of 16. For data augmentation, we use the large-scale jittering (LSJ) augmentation~\cite{ghiasi2021simple,du2021simple} with a random scale sampled from range 0.1 to 2.0 followed by a fixed size crop to $1024\x1024$. We use the standard Mask R-CNN inference setting where we resize an image with shorter side to 800 and longer side up-to 1333. We also report FLOPs and fps. FLOPs are averaged over 100 validation images (COCO images have varying sizes). Frames-per-second (fps) is measured on a V100 GPU with a batch size of 1 by taking the average runtime on the entire validation set including post-processing time.

\noindent\textbf{Semantic segmentation.} We follow the same settings as~\cite{cheng2021maskformer} to train our models, except: 1) a learning rate multiplier of 0.1 is applied to \emph{both} CNN and Transformer backbones instead of only applying it to CNN backbones in~\cite{cheng2021maskformer},
2) both ResNet and Swin backbones use an initial learning rate of $0.0001$ and a weight decay of $0.05$, instead of using different learning rates in~\cite{cheng2021maskformer}.

\subsection{Main results}

\noindent\textbf{Panoptic segmentation.} We compare \modelname with state-of-the-art models for panoptic segmentation on the COCO panoptic~\cite{kirillov2017panoptic} dataset in \tabref{tab:panseg:coco}. \modelname consistently outperforms MaskFormer by more than 5 PQ across different backbones while converging $6\x$ faster. With Swin-L backbone,
our \modelname sets a new state-of-the-art of 57.8 PQ, outperforming existing state-of-the-art~\cite{cheng2021maskformer} by 5.1 PQ and concurrent work, K-Net~\cite{zhang2021knet}, by 3.2 PQ.
\modelname even outperforms the best ensemble models with extra training data in the COCO challenge (see \appref{app:results:panoptic} for test set results).

Beyond the PQ metric, our \modelname also achieves higher performance on two other metrics compared to DETR~\cite{detr} and MaskFormer: AP$^\text{Th}_\text{pan}$, which is the AP evaluated on the 80 ``thing'' categories using \emph{instance segmentation annotation}, and mIoU$_\text{pan}$, which is the mIoU evaluated on the 133 categories for semantic segmentation converted from panoptic segmentation annotation. This shows \modelname's universality: %
trained \emph{only} with panoptic segmentation annotations, it can be used for instance and semantic segmentation.

\begin{table}[t]
  \centering
  \tablestyle{2pt}{1.2}
  \scriptsize
  \begin{tabular}{l | lc | x{35}x{35}}
  method & backbone & crop size & mIoU (s.s.) & mIoU (m.s.) \\
  \shline
  MaskFormer~\cite{cheng2021maskformer} & R50 & $512$ & 44.5 & 46.7 \\
  \textbf{\modelname} (ours) & R50 & $512$ & \textbf{47.2} & \textbf{49.2} \\
  \hline\hline
  \demph{Swin-UperNet~\cite{liu2021swin,xiao2018unified}} & \demph{Swin-T\phantom{$^{\text{\textdagger}}$}} & \demph{$512$} & \demph{-} & \demph{46.1} \\
  MaskFormer~\cite{cheng2021maskformer} & Swin-T\phantom{$^{\text{\textdagger}}$} & $512$ & 46.7 & 48.8 \\
  \textbf{\modelname} (ours) & Swin-T & $512$ & \textbf{47.7} & \textbf{49.6} \\
  \hline\hline
  MaskFormer~\cite{cheng2021maskformer} & Swin-L$^{\text{\textdagger}}$ & $640$ & 54.1 & 55.6 \\
  FaPN-MaskFormer~\cite{fapn,cheng2021maskformer} & Swin-L-FaPN$^{\text{\textdagger}}$ & $640$ & 55.2 & 56.7 \\
  BEiT-UperNet~\cite{beit,xiao2018unified} & BEiT-L$^{\text{\textdagger}}$ & $640$ & - & 57.0 \\
  \hline
  \multirow{2}{*}{\textbf{\modelname} (ours)}
  & Swin-L$^{\text{\textdagger}}$ & $640$ & 56.1 & 57.3 \\
  & Swin-L-FaPN$^{\text{\textdagger}}$ & $640$ & \textbf{56.4} & \textbf{57.7} \\
  \end{tabular}

  \caption{\textbf{Semantic segmentation on ADE20K \texttt{val} with 150 categories.} \modelname consistently outperforms MaskFormer~\cite{cheng2021maskformer} by a large margin with different backbones (all \modelname models use MSDeformAttn~\cite{zhu2021deformable} as pixel decoder, except Swin-L-FaPN uses FaPN~\cite{fapn}). Our best model outperforms the best specialized model, BEiT~\cite{beit}. We report both single-scale (s.s.) and multi-scale (m.s.) inference results.
  Backbones pre-trained on ImageNet-22K are marked with $^{\text{\textdagger}}$.}
  \vspace{-3mm}

\label{tab:semseg:ade20k}
\end{table}

\begin{table*}[t]
  \begin{subtable}{0.49\linewidth}
  \centering
  \tablestyle{3pt}{1.2}
  \scriptsize
  \begin{tabular}{y{80} | x{30}x{30}x{30} | x{30}}
   & AP & PQ & mIoU & FLOPs \\
  \shline
  \textbf{\modelname} (ours) & \textbf{43.7} \phantom{\dt{-0.0}} & \textbf{51.9} \phantom{\dt{-0.0}} & \textbf{47.2} \phantom{\dt{-0.0}} & 226G \\
  \hline
  $-$ masked attention & 37.8 \dt{-5.9} & 47.1 \dt{-4.8} & 45.5 \dt{-1.7} & 213G \\
  $-$ high-resolution features & 41.5 \dt{-2.2} & 50.2 \dt{-1.7} & 46.1 \dt{-1.1} & 218G \\
  \multicolumn{5}{c}{~}\\
  \multicolumn{5}{c}{~}\\
  \end{tabular}
  \caption{Masked attention and high-resolution features (from efficient multi-scale strategy) lead to the most gains. More detailed ablations are in \tabref{tab:ablation:maskformer:a} and \tabref{tab:ablation:maskformer:b}. We remove one component at a time.
  }
  \label{tab:ablation:transformer:a}
  \end{subtable}\hspace{2mm}
  \begin{subtable}{0.49\linewidth}
  \centering
  \tablestyle{3pt}{1.2}
  \scriptsize
  \begin{tabular}{y{80} | x{30}x{30}x{30} | x{30}}
  & AP & PQ & mIoU & FLOPs \\
  \shline
  \modelname (ours) & \textbf{43.7} \phantom{\dt{-0.0}} & \textbf{51.9} \phantom{\dt{-0.0}} & \textbf{47.2} \phantom{\dt{-0.0}} & 226G \\
  \hline
  $-$ learnable query features & 42.9 \dt{-0.8} & 51.2 \dt{-0.7} & 45.4 \dt{-1.8} & 226G \\
  $-$ cross-attention first & 43.2 \dt{-0.5} & 51.6 \dt{-0.3} & 46.3 \dt{-0.9} & 226G \\
  $-$ remove dropout & 43.0 \dt{-0.7} & 51.3 \dt{-0.6} & 47.2 \dt{-0.0} & 226G \\
  \hline
  $-$ all 3 components above & 42.3 \dt{-1.4} & 50.8 \dt{-1.1} & 46.3 \dt{-0.9} & 226G \\
  \end{tabular}
  \caption{Optimization improvements increase the performance without introducing extra compute. Following DETR~\cite{detr}, query features are zero-initialized when not learnable. We remove one component at a time.}
  \label{tab:ablation:transformer:b}
  \end{subtable}\vspace{2mm}
  \begin{subtable}{0.3\linewidth}
  \centering
  \tablestyle{1pt}{1.2}
  \scriptsize
  \begin{tabular}{l | x{20}x{20}x{20} | x{24}}
   & AP & PQ & mIoU & FLOPs \\
  \shline
  cross-attention & 37.8 & 47.1 & 45.5 & 213G \\
  \hline
  SMCA~\cite{gao2021smca} & 37.9 & 47.2 & \underline{46.6} & 213G \\
  mask pooling~\cite{zhang2021knet} & \underline{43.1} & \underline{51.5} & 46.0 & 217G \\
  \hline
  \textbf{masked attention} & \textbf{43.7} & \textbf{51.9} & \textbf{47.2} & 226G \\
  \multicolumn{5}{c}{~}\\
  \end{tabular}
  \caption{\textbf{Masked attention.} Our masked attention performs better than other variants of cross-attention across all tasks.
  }
  \label{tab:ablation:maskformer:a}
  \end{subtable}\hspace{2mm}
  \begin{subtable}{0.35\linewidth}
  \centering
  \tablestyle{1pt}{1.2}
  \scriptsize
  \begin{tabular}{l | x{20}x{20}x{20} | x{24}}
   & AP & PQ & mIoU & FLOPs \\
  \shline
  single scale ($1/32$) & 41.5 & 50.2 & 46.1 & 218G \\
  single scale ($1/16$) & 43.0 & 51.5 & 46.5 & 222G \\
  single scale ($1/8$) & \textbf{44.0} & \underline{51.8} & \textbf{47.4} & 239G \\
  \hline
  na\"ive m.s.\ (3 scales) & \textbf{44.0} & \textbf{51.9} & 46.3 & 247G \\
  \hline
  \textbf{efficient m.s.} (3 scales) & \underline{43.7} & \textbf{51.9} & \underline{47.2} & 226G \\
  \end{tabular}
  \caption{\textbf{Feature resolution.}
  High-resolution features (single scale $1/8$) are important. Our efficient multi-scale (efficient m.s.) strategy effectively reduces the FLOPs.
  }
  \label{tab:ablation:maskformer:b}
  \end{subtable}\hspace{2mm}
  \begin{subtable}{0.32\linewidth}
  \centering
  \tablestyle{1pt}{1.2}
  \scriptsize
  \begin{tabular}{l | x{20}x{20}x{20} | x{24}}
   & AP & PQ & mIoU & FLOPs \\
  \shline
  FPN~\cite{lin2016feature} & 41.5 & 50.7 & 45.6 & 195G \\
  \hline
  Semantic FPN~\cite{kirillov2019panopticfpn} & 42.1 & 51.2 & 46.2 & 258G \\
  FaPN~\cite{fapn} & 42.4 & \underline{51.8} & \underline{46.8} & - \\
  BiFPN~\cite{tan2020efficientdet} & \underline{43.5} & \underline{51.8} & 45.6 & 204G \\
  \hline
  \textbf{MSDeformAttn}~\cite{zhu2021deformable} & \textbf{43.7} & \textbf{51.9} & \textbf{47.2} & 226G \\
  \end{tabular}
  \caption{\textbf{Pixel decoder.} MSDeformAttn~\cite{zhu2021deformable} consistently performs the best across all tasks.
  \phantom{pad} \phantom{pad} \phantom{pad} \phantom{pad} \phantom{pad} \phantom{pad} \phantom{pad} \phantom{pad} \phantom{pad} %
  }
  \label{tab:ablation:maskformer:c}
  \end{subtable}
  \caption{\textbf{\modelname ablations.} We perform ablations on three tasks: instance (AP on COCO \texttt{val2017}), panoptic (PQ on COCO panoptic \texttt{val2017}) and semantic (mIoU on ADE20K \texttt{val}) segmentation. FLOPs are measured on COCO instance segmentation.
  }
  \label{tab:ablation:maskformer}
  \vspace{-3mm}
\end{table*}

\noindent\textbf{Instance segmentation.} We compare \modelname with state-of-the-art models %
on the COCO~\cite{lin2014coco} dataset in \tabref{tab:insseg:coco}. With ResNet~\cite{he2016deep} backbone, \modelname outperforms a strong Mask R-CNN~\cite{he2017mask} baseline using large-scale jittering (LSJ) augmentation~\cite{ghiasi2021simple,du2021simple} while requiring $8\x$ fewer training iterations.
With Swin-L backbone, \modelname outperforms the state-of-the-art HTC++~\cite{chen2019hybrid}.
Although we only observe $+0.6$ AP improvement over HTC++, the Boundary AP~\cite{cheng2021boundary} improves by $2.1$, suggesting that our predictions have a better boundary quality %
thanks to the high-resolution mask predictions.
Note that for a fair comparison, we only consider single-scale inference and models trained with only COCO \texttt{train2017} set data.

With a ResNet-50 backbone \modelname  improves over MaskFormer on small objects by 7.0 AP$^\text{S}$, while overall the highest gains come from large objects ($+10.6$ AP$^\text{L}$). The performance on AP$^\text{S}$ still lags behind other state-of-the-art models.
Hence there still remains room for improvement on small objects, \eg, by using dilated backbones like in DETR~\cite{detr}, which we leave for future work.

\noindent\textbf{Semantic segmentation.} We compare \modelname with state-of-the-art models for semantic segmentation on the ADE20K~\cite{zhou2017ade20k} dataset in \tabref{tab:semseg:ade20k}. \modelname outperforms MaskFormer~\cite{cheng2021maskformer} across different backbones, suggesting that the proposed improvements even boost semantic segmentation results where~\cite{cheng2021maskformer} was already state-of-the-art. With Swin-L as  backbone and FaPN~\cite{fapn} as  pixel decoder, \modelname sets a new state-of-the-art of 57.7 mIoU. We also report the test set results in \appref{app:results:semantic}.

\subsection{Ablation studies}
We now analyze \modelname through a series of ablation studies using a ResNet-50 backbone~\cite{he2016deep}. To test the generality of the proposed components  for universal image segmentation, \emph{all ablations are performed on three tasks}.

\noindent\textbf{Transformer decoder.}
We validate the importance of each component %
by removing them one at a time. As shown in \tabref{tab:ablation:transformer:a}, masked attention leads to the biggest improvement across all tasks. The improvement is larger for %
instance and panoptic segmentation
than for semantic segmentation. Moreover, using high-resolution features from the efficient multi-scale strategy is also important. \tabref{tab:ablation:transformer:b} shows additional optimization improvements further improve the performance without extra computation.

\noindent\textbf{Masked attention.}
Concurrent work has proposed other variants of cross-attention~\cite{gao2021smca,meng2021cdetr} that aim to improve the convergence and performance of DETR~\cite{detr} for object detection. Most recently, K-Net~\cite{zhang2021knet} replaced cross-attention with a mask pooling operation that averages features within mask regions. We validate the importance of our masked attention in \tabref{tab:ablation:maskformer:a}. While existing cross-attention variants may improve on a specific task, our masked attention performs the best on all three tasks.

\noindent\textbf{Feature resolution.}
\tabref{tab:ablation:maskformer:b} shows that \modelname benefits from using high-resolution features (\eg, a single scale of $1/8$) in the Transformer decoder. However, this introduces additional computation. Our efficient multi-scale (efficient m.s.) strategy effectively reduces the FLOPs without affecting the performance. Note that, naively concatenating multi-scale features as input to every Transformer decoder layer (na\"ive m.s.) does not yield additional gains.

\noindent\textbf{Pixel decoder.}
As shown in \tabref{tab:ablation:maskformer:c}, \modelname is compatible with any existing pixel decoder. However, we observe different pixel decoders specialize in different tasks: while BiFPN~\cite{tan2020efficientdet} performs better on instance-level segmentation, FaPN~\cite{fapn} works better for semantic segmentation. Among all studied pixel decoders, the MSDeformaAttn~\cite{zhu2021deformable} consistently performs the best across all tasks and thus is selected as our default. This set of ablations also suggests that designing a module like a pixel decoder for a specific task does not guarantee generalization across  segmentation tasks. \modelname, as a universal model, could serve as a testbed for a generalizable module design.

\begin{table}[t]
  \centering
  \tablestyle{3pt}{1.2}
  \scriptsize
  \begin{tabular}{x{40}x{40}| x{30}x{30}x{30} | x{24}}
  \phantom{matching loss} matching loss & \phantom{matching loss} training loss & AP (COCO) & PQ (COCO) & mIoU (ADE20K) & memory (COCO) \\
  \shline
  \multirow{2}{*}{mask \phantom{(ours)}} & mask \phantom{(ours)} & 41.0 & 50.3 & 45.9 & 18G \\ %
   & point \phantom{(ours)} & 41.0 & 50.8 & 45.9 & \phantom{0}6G \\ %
  \hline
  \multirow{2}{*}{\textbf{point} (ours)} & mask \phantom{(ours)} & 43.1 & 51.4 & \textbf{47.3} & 18G \\ %
   & \textbf{point} (ours) & \textbf{43.7} & \textbf{51.9} & 47.2 & \phantom{0}6G \\ %
  \end{tabular}
  \caption{\textbf{Calculating loss with points \vs masks.} Training with point loss reduces training memory without influencing the performance. Matching with point loss further improves performance.}
  \label{tab:ablation:matching}
  \vspace{-3mm}
\end{table}

\noindent\textbf{Calculating loss with points \vs masks.}
In \tabref{tab:ablation:matching} we study the performance and memory implications when calculating the loss based on either mask or sampled points. Calculating the final training loss with sampled points reduces training memory by $3\x$ without affecting the performance. Additionally, calculating the matching loss with sampled points improves performance across all three tasks.

\begin{figure}[t]
    \begin{subtable}{1.0\linewidth}
    \centering
    \bgroup
    \def\arraystretch{0.2}
    \setlength\tabcolsep{0.2pt}
    \begin{tabular}{cccc}
    \includegraphics[width=0.24\linewidth]{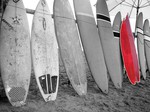} &
    \includegraphics[width=0.24\linewidth]{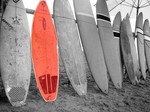} &
    \includegraphics[width=0.24\linewidth]{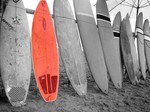} &
    \includegraphics[width=0.24\linewidth]{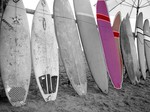} \\
    \includegraphics[width=0.24\linewidth]{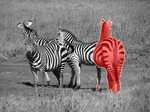} &
    \includegraphics[width=0.24\linewidth]{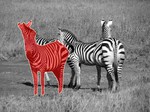} &
    \includegraphics[width=0.24\linewidth]{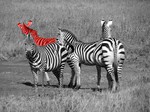} &
    \includegraphics[width=0.24\linewidth]{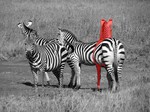} \\
    \includegraphics[width=0.24\linewidth]{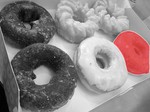} &
    \includegraphics[width=0.24\linewidth]{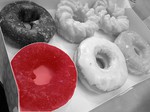} &
    \includegraphics[width=0.24\linewidth]{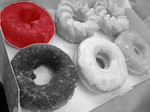} &
    \includegraphics[width=0.24\linewidth]{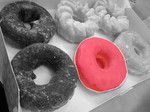} \\
    \end{tabular} \egroup
    \end{subtable}\vspace{0mm}
    \begin{subtable}{0.48\linewidth}
      \centering
      \tablestyle{3pt}{1.2}
      \scriptsize
      \begin{tabular}{c}
       \includegraphics[width=0.99\linewidth]{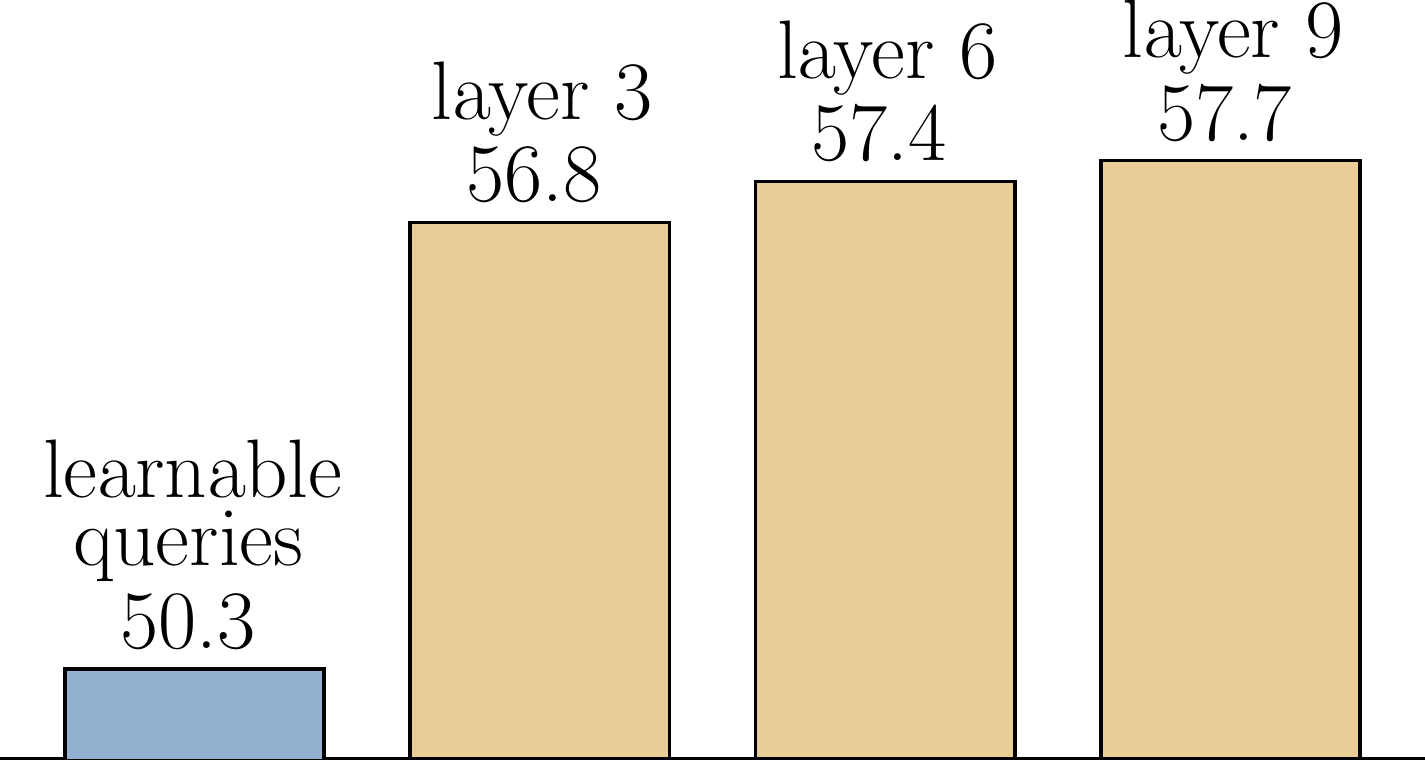} \\
      \end{tabular}
      \caption*{AR@100 on COCO \texttt{val2017}}
    \end{subtable}
    \begin{subtable}{0.48\linewidth}
      \centering
      \tablestyle{3pt}{1.2}
      \scriptsize
      \begin{tabular}{c}
       \includegraphics[width=0.99\linewidth]{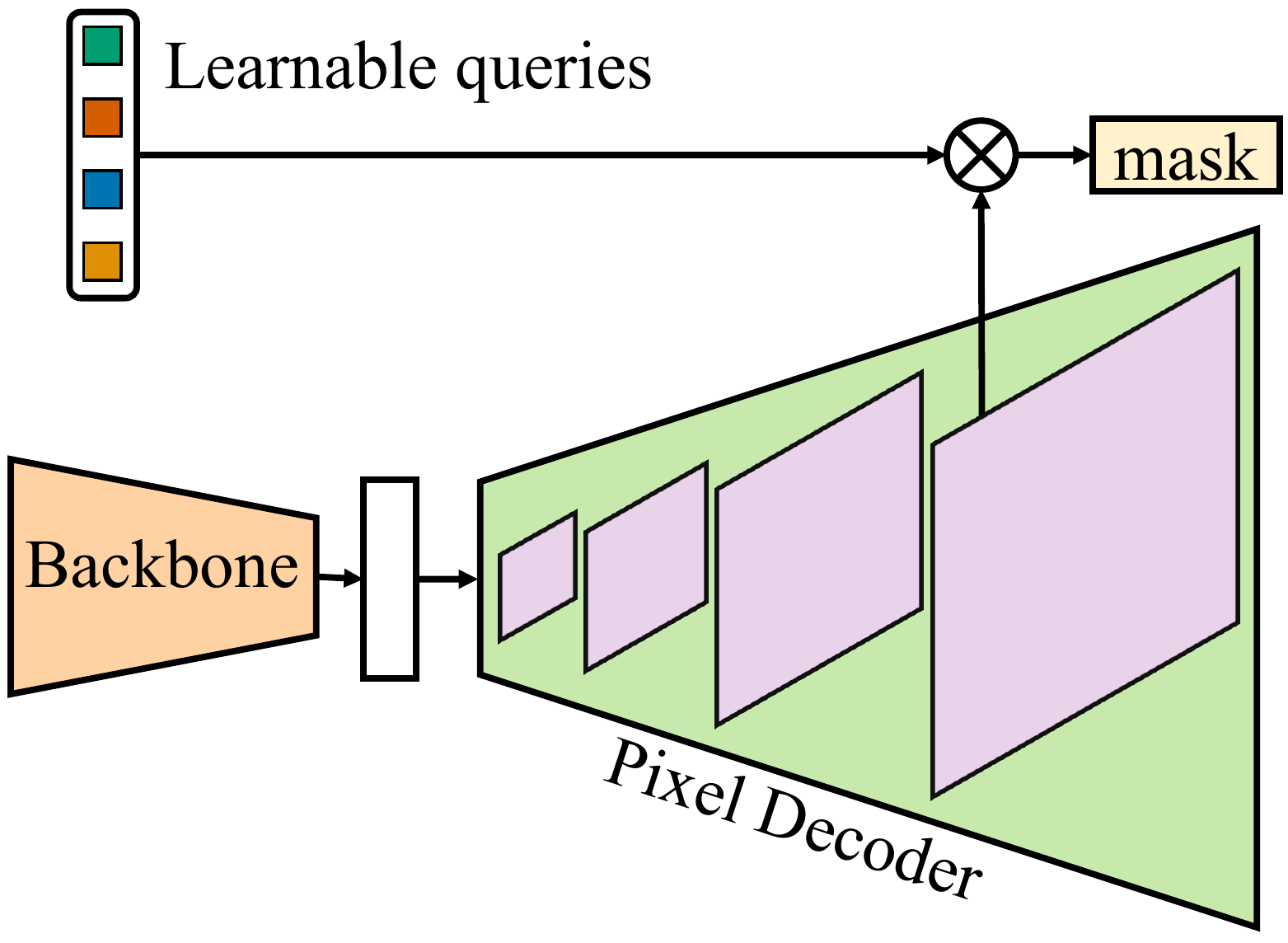} \\
      \end{tabular}
    \end{subtable}
  \vspace{-2mm}
  \caption{
  \textbf{Learnable queries as ``region proposals''.} \emph{Top:} We visualize mask predictions of four selected learnable queries \emph{before} feeding them into the Transformer decoder (using R50 backbone).
  \emph{Bottom left:} We calculate the class-agnostic average recall with 100 proposals (AR@100) and observe that these learnable queries provide good proposals compared to the final predictions of \modelname after the Transformer decoder layers (layer 9).
  \emph{Bottom right:} Illustration of proposal generation process.
  }
  \label{fig:analysis:query}
\end{figure}

\noindent\textbf{Learnable queries as region proposals.} Region proposals~\cite{uijlings2013selective,arbelaez2014multiscale}, either in the form of boxes or masks, are regions that are likely to be ``objects.'' With learnable queries being supervised by the mask loss, predictions from learnable queries can serve as mask proposals. In \figref{fig:analysis:query} top, we visualize mask predictions of selected learnable queries \emph{before} feeding them into the Transformer decoder (the proposal generation process is shown in \figref{fig:analysis:query} bottom right). In \figref{fig:analysis:query} bottom left, we further perform a quantitative analysis on the quality of these proposals by calculating the class-agnostic average recall with 100 predictions (AR@100) on COCO \texttt{val2017}. We find these learnable queries already achieve good AR@100 compared to the final predictions of \modelname after the Transformer decoder layers, \ie, layer 9, and AR@100 consistently improves with more decoder layers.

\begin{table}[t]
  \centering

  \tablestyle{1pt}{1.2}\scriptsize\begin{tabular}{l|l | x{16}x{20}x{24} |x{36}x{16}}
  & & \multicolumn{3}{c|}{panoptic model} & \multicolumn{2}{c}{semantic model} \\
  method & backbone & PQ & AP$^\text{Th}_\text{pan}$ & mIoU$_\text{pan}$ & mIoU (s.s.) & (m.s.) \\
  \shline
  Panoptic FCN~\cite{li2021fully} & Swin-L$^{\text{\textdagger}}$ & 65.9\phantom{$^*$} & - & - & - & - \\
  Panoptic-DeepLab~\cite{cheng2020panoptic} & SWideRNet~\cite{chen2020scaling} & 66.4\phantom{$^*$} & 40.1\phantom{$^*$} & 82.2\phantom{$^*$} & - & - \\
  \demph{Panoptic-DeepLab~\cite{cheng2020panoptic}} & \demph{SWideRNet~\cite{chen2020scaling}} & \demph{67.5$^*$} & \demph{43.9$^*$} & \demph{82.9$^*$} & - & - \\
  \hline
  SETR~\cite{zheng2021rethinking} & ViT-L$^{\text{\textdagger}}$~\cite{dosovitskiy2020vit} & - & - & - & - & 82.2 \\
  SegFormer~\cite{xie2021segformer} & MiT-B5~\cite{xie2021segformer} & - & - & - & - & 84.0 \\
  \hline\hline
  \multirow{3}{*}{\textbf{\modelname} (ours)}
  & R50\phantom{$^{\text{\textdagger}}$} & 62.1\phantom{$^*$} & 37.3\phantom{$^*$} & 77.5\phantom{$^*$} & 79.4 & 82.2 \\
  & Swin-B$^{\text{\textdagger}}$ & 66.1\phantom{$^*$} & 42.8\phantom{$^*$} & 82.7\phantom{$^*$} & \textbf{83.3} & \textbf{84.5} \\
  & Swin-L$^{\text{\textdagger}}$ & \textbf{66.6}\phantom{$^*$} & \textbf{43.6}\phantom{$^*$} & \textbf{82.9}\phantom{$^*$} & \textbf{83.3} & 84.3 \\
  \end{tabular}
  \vspace{-2mm}

   \caption{\textbf{Cityscapes \texttt{val}.} \modelname is competitive to specialized models on Cityscapes. Panoptic segmentation models use single-scale inference by default, multi-scale numbers are marked with~$^*$. For semantic segmentation, we report both single-scale (s.s.) and multi-scale (m.s.) inference results. Backbones pre-trained on ImageNet-22K are marked with $^{\text{\textdagger}}$.}
  \vspace{-3mm}

\label{tab:benchmark:cityscapes}
\end{table}

\subsection{Generalization to other datasets}
To show our \modelname can generalize beyond the COCO dataset, we further perform experiments on other popular image segmentation datasets. In \tabref{tab:benchmark:cityscapes}, we show results on Cityscapes~\cite{Cordts2016Cityscapes}. Please see \appref{app:datasets} for detailed training settings on each dataset as well as more results on  ADE20K~\cite{zhou2017ade20k} and Mapillary Vistas~\cite{neuhold2017mapillary}.

We observe that our \modelname is competitive to state-of-the-art methods on these datasets as well. It suggests \modelname can serve as a universal image segmentation model and results generalize across datasets.

\begin{table}[t]
\centering
  \begin{subtable}{0.4\linewidth}
  \centering
  \tablestyle{2pt}{1.2}
  \scriptsize
  \begin{tabular}{l | x{16}x{16}x{16}}
   & PQ & AP & mIoU \\
  \shline
  panoptic & \demph{51.9} & 41.7 & \textbf{61.7} \\
  \hline
  instance & - & \textbf{43.7} & - \\
  semantic & - & - & 61.5 \\
  \end{tabular}
  \caption{COCO}
  \label{tab:analysis:universal:coco}
  \end{subtable}
  \begin{subtable}{0.25\linewidth}
  \centering
  \tablestyle{1pt}{1.2}
  \scriptsize
  \begin{tabular}{x{16}x{16}x{16}}
  PQ & AP & mIoU \\
  \shline
  \demph{39.7} & \textbf{26.5} & 46.1 \\
  \hline
  - & 26.4 & - \\
  - & - & \textbf{47.2} \\
  \end{tabular}
  \caption{ADE20K}
  \label{tab:analysis:universal:ade20k}
  \end{subtable}
  \begin{subtable}{0.25\linewidth}
  \centering
  \tablestyle{1pt}{1.2}
  \scriptsize
  \begin{tabular}{x{16}x{16}x{16}}
  PQ & AP & mIoU \\
  \shline
  \demph{62.1} & 37.3 & 77.5 \\
  \hline
  - & \textbf{37.4} & - \\
  - & - & \textbf{79.4} \\
  \end{tabular}
  \caption{Cityscapes}
  \label{tab:analysis:universal:cs}
  \end{subtable}
  \vspace{-2mm}
  \caption{\textbf{Limitations of \modelname.} Although a single \modelname can address any segmentation task, we still need to train it on different tasks. Across three datasets we find \modelname trained with panoptic annotations performs slightly worse than the exact same model trained specifically for instance and semantic segmentation tasks with the corresponding data.}
  \label{tab:analysis:universal}
  \vspace{-3mm}
\end{table}

\subsection{Limitations}
Our ultimate goal is to train a {\em single} model for \emph{all} image segmentation tasks. In \tabref{tab:analysis:universal}, we find \modelname trained on panoptic segmentation only performs slightly worse than the exact same model trained with the corresponding annotations for instance and semantic segmentation tasks across three datasets.
This suggests that even though \modelname can generalize to different tasks, it still needs to be trained for those specific tasks.
In the future, we hope to develop a model that can be trained only once for multiple tasks and even for multiple datasets.

Furthermore, as seen in Tables~\ref{tab:insseg:coco} and \ref{tab:ablation:maskformer:b}, even though it improves over baselines, \modelname struggles with segmenting small objects and is unable to fully leverage multi-scale features. We believe better utilization of the feature pyramid and designing losses for small objects are critical.

%% file: sections/concl.tex
\section{Conclusion}
We present \modelname for universal image segmentation. Built upon a simple meta framework~\cite{cheng2021maskformer} with a new Transformer decoder using the proposed masked attention, \modelname obtains top results in all three major image segmentation tasks (panoptic, instance and semantic) on four popular datasets, outperforming even the best specialized models designed for each benchmark while remaining easy to train. \modelname saves $3\x$ research effort compared to designing specialized models for each task, and it is accessible to users with limited computational resources. We hope to attract interest in universal model design. %

%% file: sections/appendix.tex
We first provide more results for \modelname with different backbones as well as test-set performance on  standard benchmarks (\appref{app:results}): We use COCO panoptic~\cite{kirillov2017panoptic} for panoptic, COCO~\cite{lin2014coco} for instance, and ADE20K~\cite{zhou2017ade20k} for semantic segmentation. Then, we provide more detailed results on additional datasets (\appref{app:datasets}). Finally, we provide additional ablation studies (\appref{app:abl}) and visualization of \modelname predictions for all three segmentation tasks (\appref{app:vis}).

\begin{table*}[t]
  \centering

  \tablestyle{4pt}{1.2}\scriptsize\begin{tabular}{c|l | lcc | x{20}x{20}x{20} | x{20}x{20} |x{24}x{24}}
  & method & backbone & search space & epochs & PQ & PQ$^\text{Th}$ & PQ$^\text{St}$ & AP$^\text{Th}_\text{pan}$ & mIoU$_\text{pan}$ & \#params. & FLOPs \\
  \shline
  \multirow{8}{*}{\rotatebox{90}{CNN backbones}}
  & \multirow{2}{*}{DETR~\cite{detr}}
  & R50 & 100 queries & 500+25 & 43.4 & 48.2 & 36.3 & 31.1 & - & - & - \\
  & & R101 & 100 queries & 500+25 & 45.1 & 50.5 & 37.0 & 33.0 & - & - & - \\
  \cline{2-12}
  & K-Net~\cite{zhang2021knet}
  & R50 & 100 queries & 36 & 47.1 & 51.7 & 40.3 & - & - & - & - \\
  & Panoptic SegFormer~\cite{li2021panopticsegformer}
  & R50 & 400 queries & 50 & 50.0 & 56.1 & 40.8 & - & - & 47M & \phantom{0}246G \\
  \cline{2-12}
  & \multirow{2}{*}{MaskFormer~\cite{cheng2021maskformer}}
  & R50 & 100 queries & 300 & 46.5 & 51.0 & 39.8 & 33.0 & 57.8 & \phantom{0}45M & \phantom{0}181G \\
  & & R101 & 100 queries & 300 & 47.6 & 52.5 & 40.3 & 34.1 & 59.3 & \phantom{0}64M & \phantom{0}248G \\
  \cline{2-12}
  & \multirow{2}{*}{\textbf{\modelname} (ours)}
  & R50 & 100 queries & 50 & 51.9 & 57.7 & 43.0 & 41.7 & 61.7 & \phantom{0}44M & \phantom{0}226G \\
  & & R101 & 100 queries & 50 & \textbf{52.6} & \textbf{58.5} & \textbf{43.7} & \textbf{42.6} & \textbf{62.4} & \phantom{0}63M & \phantom{0}293G \\
  \hline\hline
  \multirow{14}{*}{\rotatebox{90}{Transformer backbones}}
  & \multirow{2}{*}{Max-DeepLab~\cite{wang2021max}}
  & Max-S & 128 queries & 216 & 48.4 & 53.0 & 41.5 & - & - & \phantom{0}62M & \phantom{0}324G \\
  & & Max-L & 128 queries & 216 & 51.1 & 57.0 & 42.2 & - & - & 451M & 3692G \\
  \cline{2-12}
  & Panoptic SegFormer~\cite{li2021panopticsegformer}
  & PVTv2-B5~\cite{wang2021pvtv2} & 400 queries & 50 & 54.1 & 60.4 & 44.6 & - & - & 101M & \phantom{0}391G \\
  & K-Net~\cite{zhang2021knet}
  & Swin-L$^{\text{\textdagger}}$ & 100 queries & 36 & 54.6 & 60.2 & 46.0 & - & - & - & - \\
  \cline{2-12}
  & \multirow{5}{*}{MaskFormer~\cite{cheng2021maskformer}}
  & Swin-T & 100 queries & 300 & 47.7 & 51.7 & 41.7 & 33.6 & 60.4 & \phantom{0}42M & \phantom{0}179G \\
  & & Swin-S & 100 queries & 300 & 49.7 & 54.4 & 42.6 & 36.1 & 61.3 & \phantom{0}63M & \phantom{0}259G \\
  & & Swin-B & 100 queries & 300 & 51.1 & 56.3 & 43.2 & 37.8 & 62.6 & 102M & \phantom{0}411G \\
  & & Swin-B$^{\text{\textdagger}}$ & 100 queries & 300 & 51.8 & 56.9 & 44.1 & 38.5 & 63.6 & 102M & \phantom{0}411G \\
  & & Swin-L$^{\text{\textdagger}}$ & 100 queries & 300 & 52.7 & 58.5 & 44.0 & 40.1 & 64.8 & 212M & \phantom{0}792G \\
  \cline{2-12}
  & \multirow{5}{*}{\textbf{\modelname} (ours)}
  & Swin-T & 100 queries & 50 & 53.2 & 59.3 & 44.0 & 43.3 & 63.2 & \phantom{0}47M & \phantom{0}232G \\
  & & Swin-S & 100 queries & 50 & 54.6 & 60.6 & 45.7 & 44.7 & 64.2 & \phantom{0}69M & \phantom{0}313G \\
  & & Swin-B & 100 queries & 50 & 55.1 & 61.0 & 46.1 & 45.2 & 65.1 & 107M & \phantom{0}466G \\
  & & Swin-B$^{\text{\textdagger}}$ & 100 queries & 50 & 56.4 & 62.4 & 47.3 & 46.3 & 67.1 & 107M & \phantom{0}466G \\
  & & Swin-L$^{\text{\textdagger}}$ & \emph{200 queries} & \emph{100} & \textbf{57.8} & \textbf{64.2} & \textbf{48.1} & \textbf{48.6} & \textbf{67.4} & 216M & \phantom{0}868G \\
  \end{tabular}

  \caption{\textbf{Panoptic segmentation on COCO panoptic \texttt{val2017} with 133 categories.} \modelname outperforms \emph{all} existing panoptic segmentation models by a large margin with different backbones on all metrics. Our best model sets a new state-of-the-art of $57.8$ PQ.
   Besides PQ for panoptic segmentation, we also report AP$^\text{Th}_\text{pan}$ (the AP evaluated on the 80 ``thing'' categories using \emph{instance segmentation annotation}) and mIoU$_\text{pan}$ (the mIoU evaluated on the 133 categories for semantic segmentation converted from panoptic segmentation annotation) of the same model trained for panoptic segmentation (\textbf{note: we train all our models with panoptic segmentation annotation only}).
Backbones pre-trained on ImageNet-22K are marked with $^{\text{\textdagger}}$.
}
\label{tab:panseg:coco_full}
\end{table*}

\begin{table*}[t]
  \centering

  \tablestyle{5pt}{1.2}\scriptsize\begin{tabular}{ll | x{20}x{20}x{20} | x{20}x{20}}
  method & backbone & PQ & PQ$^\text{Th}$ & PQ$^\text{St}$ & SQ & RQ \\
  \shline
  Max-DeepLab~\cite{wang2021max} & Max-L & 51.3 & 57.2 & 42.4 & 82.5 & 61.3 \\
  Panoptic FCN~\cite{li2021fully} & Swin-L & 52.7 & 59.4 & 42.5 & - & - \\
  MaskFormer~\cite{cheng2021maskformer} & Swin-L & 53.3 & 59.1 & 44.5 & 82.0 & 64.1 \\
  Panoptic SegFormer~\cite{li2021panopticsegformer} & PVTv2-B5~\cite{wang2021pvtv2} & 54.4 & 61.1 & 44.3 & 83.3 & 64.6 \\
  K-Net~\cite{zhang2021knet} & Swin-L & 55.2 & 61.2 & 46.2 & - & - \\
  \hline
  Megvii (challenge winner) & - & 54.7 & 64.6 & 39.8 & 83.6 & 64.3 \\
  \hline
  \textbf{\modelname} (ours) & Swin-L & \textbf{58.3} & \textbf{65.1} & \textbf{48.1} & \textbf{84.1} & \textbf{68.6} \\
  \end{tabular}

  \caption{\textbf{Panoptic segmentation on COCO panoptic \texttt{test-dev} with 133 categories.} \modelname, without any bells-and-whistles, outperforms the challenge winner (which uses extra training data, model ensemble, \etc) on the \texttt{test-dev} set. We only train our model on the COCO \texttt{train2017} set with ImageNet-22K pre-trained checkpoint.
}

\label{tab:panseg:coco_test_dev}
\end{table*}

\begin{table*}[t]
  \centering

  \tablestyle{5pt}{1.2}\scriptsize\begin{tabular}{c|l| lcc | x{20}x{20}x{20}x{20} | x{30} |x{24}x{24}}
  & method & backbone & search space & epochs & AP & AP$^\text{S}$ & AP$^\text{M}$ & AP$^\text{L}$ & AP$^\text{boundary}$ & \#params. & FLOPs \\
  \shline
  \multirow{6}{*}{\rotatebox{90}{CNN backbones}}
  & \multirow{4}{*}{Mask R-CNN~\cite{he2017mask}}
  & \demph{R50} & \demph{dense anchors} & \demph{36} & \demph{37.2} & \demph{18.6} & \demph{39.5} & \demph{53.3} & \demph{23.1} & \demph{\phantom{0}44M} & \phantom{0}\demph{201G} \\
  & & R50 & dense anchors & 400 & 42.5 & 23.8 & 45.0 & 60.0 & 28.0 & \phantom{0}46M & \phantom{0}358G \\
  \cline{3-12}
  & & \demph{R101} & \demph{dense anchors} & \demph{36} & \demph{38.6} & \demph{19.5} & \demph{41.3} & \demph{55.3} & \demph{24.5} & \demph{\phantom{0}63M} & \phantom{0}\demph{266G} \\
  & & R101 & dense anchors & 400 & 43.7 & \textbf{24.6} & 46.4 & 61.8 & 29.1 & \phantom{0}65M & \phantom{0}423G \\
  \cline{2-12}
  & \multirow{2}{*}{\textbf{\modelname} (ours)}
  & R50 & 100 queries & 50 & 43.7 & 23.4 & 47.2 & 64.8 & 30.6 & \phantom{0}44M & \phantom{0}226G \\
  & & R101 & 100 queries & 50 & \textbf{44.2} & 23.8 & \textbf{47.7} & \textbf{66.7} & \textbf{31.1} & \phantom{0}63M & \phantom{0}293G \\
  \hline\hline
  \multirow{8}{*}{\rotatebox{90}{Transformer backbones}}
  & QueryInst~\cite{QueryInst} & Swin-L$^{\text{\textdagger}}$ & 300 queries & 50 & 48.9 & 30.8 & 52.6 & 68.3 & 33.5 & - & - \\
  \cline{2-12}
  & \multirow{2}{*}{Swin-HTC++~\cite{liu2021swin,chen2019hybrid}}
   & Swin-B$^{\text{\textdagger}}$ & dense anchors & 36 & 49.1 & - & - & - & - & 160M & 1043G \\
  & & Swin-L$^{\text{\textdagger}}$ & dense anchors & 72 & 49.5 & \textbf{31.0} & 52.4 & 67.2 & 34.1 & 284M & 1470G \\
  \cline{2-12}
  & \multirow{5}{*}{\textbf{\modelname} (ours)}
   & Swin-T & 100 queries & 50 & 45.0 & 24.5 & 48.3 & 67.4 & 31.8 & \phantom{0}47M & \phantom{0}232G \\
  & & Swin-S & 100 queries & 50 & 46.3 & 25.3 & 50.3 & 68.4 & 32.9 & \phantom{0}69M & \phantom{0}313G \\
  & & Swin-B & 100 queries & 50 & 46.7 & 26.1 & 50.5 & 68.8 & 33.2 & 107M & \phantom{0}466G \\
  & & Swin-B$^{\text{\textdagger}}$ & 100 queries & 50 & 48.1 & 27.8 & 52.0 & 71.1 & 34.4 & 107M & \phantom{0}466G \\
  & & Swin-L$^{\text{\textdagger}}$ & \emph{200 queries} & \emph{100} & \textbf{50.1} & 29.9 & \textbf{53.9} & \textbf{72.1} & \textbf{36.2} & 216M & \phantom{0}868G \\
  \end{tabular}

   \caption{\textbf{Instance segmentation on COCO \texttt{val2017} with 80 categories.} \modelname outperforms strong Mask R-CNN~\cite{he2017mask} baselines with $8\x$ fewer training epochs for both AP and AP$^\text{boundary}$~\cite{cheng2021boundary} metrics. Our best model is also competitive to the state-of-the-art specialized instance segmentation model on COCO and has higher boundary quality. For a fair comparison, we only consider single-scale inference and models trained using only COCO \texttt{train2017} set data.
Backbones pre-trained on ImageNet-22K are marked with $^{\text{\textdagger}}$.}

\label{tab:insseg:coco_full}
\end{table*}

\begin{table*}[t]
  \centering

  \tablestyle{5pt}{1.2}\scriptsize\begin{tabular}{ll | x{20}x{20}x{20} | x{20}x{20}x{20}}
  method & backbone & AP & AP50 & AP75 & AP$^\text{S}$ & AP$^\text{M}$ & AP$^\text{L}$ \\
  \shline
  QueryInst~\cite{QueryInst} & Swin-L & 49.1 & 74.2 & 53.8 & \textbf{31.5} & 51.8 & 63.2 \\
  Swin-HTC++~\cite{liu2021swin,chen2019hybrid} & Swin-L & 50.2 & - & - & - & - & - \\
  \demph{Swin-HTC++~\cite{liu2021swin,chen2019hybrid} (multi-scale)} & \demph{Swin-L} & \demph{51.1} & \demph{-} & \demph{-} & \demph{-} & \demph{-} & \demph{-} \\
  \hline
  \demph{Megvii (challenge winner)} & \demph{-} & \demph{53.1} & \demph{76.8} & \demph{58.6} & \demph{36.6} & \demph{56.5} & \demph{67.7} \\
  \hline
  \textbf{\modelname} (ours) & Swin-L & \textbf{50.5} & \textbf{74.9} & \textbf{54.9} & 29.1 & \textbf{53.8} & \textbf{71.2} \\
  \end{tabular}

  \caption{\textbf{Instance segmentation on COCO \texttt{test-dev} with 80 categories.} \modelname is extremely good at segmenting large objects: we can even outperform the challenge winner (which uses extra training data, model ensemble, \etc) on AP$^\text{L}$ by a large margin without any bells-and-whistles. We only train our model on the COCO \texttt{train2017} set with ImageNet-22K pre-trained checkpoint.
}

\label{tab:insseg:coco_test_dev}
\end{table*}

\begin{table*}[t]
  \centering
  \tablestyle{5pt}{1.2}
  \scriptsize
  \begin{tabular}{c|l | lc | x{40}x{44}x{28}x{28}}
  & method & backbone & crop size & mIoU (s.s.) & mIoU (m.s.) & \#params. & FLOPs \\
  \shline
  \multirow{4}{*}{\rotatebox{90}{CNN}}
  & \multirow{2}{*}{MaskFormer~\cite{cheng2021maskformer}}
  & R50 & $512 \x 512$ & 44.5 & 46.7 & \phantom{0}41M & \phantom{0}53G \\
  & & R101 & $512 \x 512$ & 45.5 & 47.2 & \phantom{0}60M & \phantom{0}73G \\
  \cline{2-8}
  & \multirow{2}{*}{\textbf{\modelname} (ours)}
  & R50 & $512 \x 512$ & 47.2 & 49.2 & \phantom{0}44M & \phantom{0}71G \\
  & & R101 & $512 \x 512$ & \textbf{47.8} & \textbf{50.1} & \phantom{0}63M & \phantom{0}90G \\
  \hline\hline
  \multirow{14}{*}{\rotatebox{90}{Transformer backbones}}
  & Swin-UperNet~\cite{liu2021swin,xiao2018unified} & Swin-L$^{\text{\textdagger}}$ & $640 \x 640$ & - & 53.5 & 234M & 647G \\
  & FaPN-MaskFormer~\cite{fapn,cheng2021maskformer} & Swin-L$^{\text{\textdagger}}$ & $640 \x 640$ & 55.2 & 56.7 & - & - \\
  & BEiT-UperNet~\cite{beit,xiao2018unified} & BEiT-L$^{\text{\textdagger}}$ & $640 \x 640$ & - & 57.0 & 502M & - \\
  \cline{2-8}
  & \multirow{5}{*}{MaskFormer~\cite{cheng2021maskformer}}
  & Swin-T & $512 \x 512$ & 46.7 & 48.8 & \phantom{0}42M & \phantom{0}55G \\
  & & Swin-S & $512 \x 512$ & 49.8 & 51.0 & \phantom{0}63M & \phantom{0}79G \\
  & & Swin-B & $640 \x 640$ & 51.1 & 52.3 & 102M & 195G \\
  & & Swin-B$^{\text{\textdagger}}$ & $640 \x 640$ & 52.7 & 53.9 & 102M & 195G \\
  & & Swin-L$^{\text{\textdagger}}$ & $640 \x 640$ & 54.1 & 55.6 & 212M & 375G \\
  \cline{2-8}
  & \multirow{6}{*}{\textbf{\modelname} (ours)}
  & Swin-T & $512 \x 512$ & 47.7 & 49.6 & \phantom{0}47M & \phantom{0}74G \\
  & & Swin-S & $512 \x 512$ & 51.3 & 52.4 & \phantom{0}69M & \phantom{0}98G \\
  & & Swin-B & $640 \x 640$ & 52.4 & 53.7 & 107M & 223G \\
  & & Swin-B$^{\text{\textdagger}}$ & $640 \x 640$ & 53.9 & 55.1 & 107M & 223G \\
  & & Swin-L$^{\text{\textdagger}}$ & $640 \x 640$ & 56.1 & 57.3 & 215M & 403G \\
  & & Swin-L-FaPN$^{\text{\textdagger}}$ & $640 \x 640$ & \textbf{56.4} & \textbf{57.7} & 217M & - \\
  \end{tabular}

  \caption{\textbf{Semantic segmentation on ADE20K \texttt{val} with 150 categories.} \modelname consistently outperforms MaskFormer~\cite{cheng2021maskformer} by a large margin with different backbones (all \modelname models use MSDeformAttn~\cite{zhu2021deformable} as pixel decoder, except Swin-L-FaPN uses FaPN~\cite{fapn}). Our best model outperforms the best specialized model, BEiT~\cite{beit}, with less than half of the parameters. We report both single-scale (s.s.) and multi-scale (m.s.) inference results.
  Backbones pre-trained on ImageNet-22K are marked with $^{\text{\textdagger}}$.}

\label{tab:semseg:ade20k_full}
\end{table*}

\begin{table*}[ht!]
  \centering

  \tablestyle{5pt}{1.2}\scriptsize\begin{tabular}{ll| x{30}x{30}x{30}}
  method & backbone & P.A. & mIoU & score \\
  \shline
  SETR~\cite{zheng2021rethinking} & ViT-L & 78.35 & 45.03 & 61.69 \\
  Swin-UperNet~\cite{liu2021swin,xiao2018unified} & Swin-L & 78.42 & 47.07 & 62.75 \\
  MaskFormer~\cite{cheng2021maskformer} & Swin-L & 79.36 & 49.67 & 64.51 \\
  \hline
  \textbf{\modelname} (ours) & Swin-L-FaPN & \textbf{79.80} & \textbf{49.72} & \textbf{64.76} \\
  \end{tabular}

  \caption{\textbf{Semantic segmentation on ADE20K \texttt{test} with 150 categories.} \modelname outperforms previous state-of-the-art methods on all three metrics: pixel accuracy (P.A.), mIoU, as well as the final test score (average of P.A. and mIoU). We train our model on the union of ADE20K \texttt{train} and \texttt{val} set with ImageNet-22K pre-trained checkpoint following~\cite{cheng2021maskformer} and use multi-scale inference.}

\label{tab:semseg:ade20k_test}
\end{table*}

\begin{table*}[t]
  \centering

  \tablestyle{4pt}{1.2}\scriptsize\begin{tabular}{l|l | x{28}x{28}x{28}x{28} | x{28}x{28} |x{36}x{36}}
  & & \multicolumn{4}{c|}{panoptic model} & \multicolumn{2}{c|}{instance model} & \multicolumn{2}{c}{semantic model} \\
  method & backbone & PQ (s.s.) & \demph{PQ (m.s.)} & AP$^\text{Th}_\text{pan}$ & mIoU$_\text{pan}$ & AP & AP50 & mIoU (s.s.) & mIoU (m.s.) \\
  \shline
  \multirow{3}{*}{Panoptic-DeepLab~\cite{cheng2020panoptic}}
  & R50 & 60.3 & - & 32.1 & 78.7 & - & - & - & - \\
  & X71~\cite{chollet2017xception} & 63.0 & \demph{64.1} & 35.3 & 80.5 & - & - & - & - \\
  & SWideRNet~\cite{chen2020scaling} & 66.4 & \demph{67.5} & 40.1 & 82.2 & - & - & - & - \\
  \cline{1-10}
  Panoptic FCN~\cite{li2021fully} & Swin-L$^{\text{\textdagger}}$ & 65.9 & - & - & - & - & - & - & - \\
  \cline{1-10}
  Segmenter~\cite{strudel2021segmenter} & ViT-L$^{\text{\textdagger}}$ & - & - & - & - & - & - & - & 81.3 \\
  SETR~\cite{zheng2021rethinking} & ViT-L$^{\text{\textdagger}}$ & - & - & - & - & - & - & - & 82.2 \\
  SegFormer~\cite{xie2021segformer} & MiT-B5 & - & - & - & - & - & - & - & 84.0 \\
  \hline\hline
  \multirow{6}{*}{\textbf{\modelname} (ours)}
  & R50\phantom{$^{\text{\textdagger}}$} & 62.1 & - & 37.3 & 77.5 & 37.4 & 61.9 & 79.4 & 82.2 \\
  & R101\phantom{$^{\text{\textdagger}}$} & 62.4 & - & 37.7 & 78.6 & 38.5 & 63.9 & 80.1 & 81.9 \\
  \cline{2-10}
  & Swin-T\phantom{$^{\text{\textdagger}}$} & 63.9 & - & 39.1 & 80.5 & 39.7 & 66.9 & 82.1 & 83.0 \\
  & Swin-S\phantom{$^{\text{\textdagger}}$} & 64.8 & - & 40.7 & 81.8 & 41.8 & 70.4 & 82.6 & 83.6 \\
  & Swin-B$^{\text{\textdagger}}$ & 66.1 & - & 42.8 & 82.7 & 42.0 & 68.8 & \textbf{83.3} & \textbf{84.5} \\
  & Swin-L$^{\text{\textdagger}}$ & \textbf{66.6} & - & \textbf{43.6} & \textbf{82.9} & \textbf{43.7} & \textbf{71.4} & \textbf{83.3} & 84.3 \\
  \end{tabular}

  \caption{\textbf{Image segmentation results on Cityscapes \texttt{val}.} We report both single-scale (s.s.) and multi-scale (m.s.) inference results for PQ and mIoU. All other metrics are evaluated with \emph{single-scale} inference. Since \modelname is an end-to-end model, we only use single-scale inference for instance-level segmentation tasks to avoid the need for further post-processing (\eg, NMS). }

\label{tab:benchmark:cityscapes_full}
\end{table*}

\begin{table*}[t]
  \centering

  \tablestyle{3pt}{1.2}\scriptsize\begin{tabular}{l|l | x{28}x{28}x{28} | x{20}x{20}x{20}x{20} |x{36}x{36}}
  & & \multicolumn{3}{c|}{panoptic model} & \multicolumn{4}{c|}{instance model} & \multicolumn{2}{c}{semantic model} \\
  method & backbone & PQ & AP$^\text{Th}_\text{pan}$ & mIoU$_\text{pan}$ & AP & AP$^\text{S}$ & AP$^\text{M}$ & AP$^\text{L}$ & mIoU (s.s.) & mIoU (m.s.) \\
  \shline
  MaskFormer~\cite{cheng2021maskformer} & R50 & 34.7\phantom{$^*$} & - & - & - & - & - & - & - & - \\
  Panoptic-DeepLab~\cite{cheng2020panoptic} & SWideRNet~\cite{chen2020scaling} & 37.9$^*$ & - & 50.0$^*$ & - & - & - & - & - & - \\
  \hline
  Swin-UperNet~\cite{liu2021swin,xiao2018unified} & Swin-L$^{\text{\textdagger}}$ & - & - & - & - & - & - & - & - & 53.5 \\
  MaskFormer~\cite{cheng2021maskformer} & Swin-L$^{\text{\textdagger}}$ & - & - & - & - & - & - & - & 54.1 & 55.6 \\
  FaPN-MaskFormer~\cite{fapn,cheng2021maskformer} & Swin-L$^{\text{\textdagger}}$ & - & - & - & - & - & - & - & 55.2 & 56.7 \\
  BEiT-UperNet~\cite{beit,xiao2018unified} & BEiT-L$^{\text{\textdagger}}$ & - & - & - & - & - & - & - & - & 57.0 \\
  \hline\hline
  \multirow{3}{*}{\textbf{\modelname} (ours)} & R50\phantom{$^{\text{\textdagger}}$}
  & 39.7\phantom{$^*$} & 26.5 & 46.1\phantom{$^*$} & 26.4 & 10.4 & 28.9 & 43.1 & 47.2 & 49.2 \\
  & Swin-L$^{\text{\textdagger}}$ & \textbf{48.1}\phantom{$^*$} & \textbf{34.2} & 54.5\phantom{$^*$} & \textbf{34.9} & \textbf{16.3} & \textbf{40.0} & \textbf{54.7} & 56.1 & 57.3 \\
  & Swin-L-FaPN$^{\text{\textdagger}}$ & 46.2\phantom{$^*$} & 33.2 & \textbf{55.4}\phantom{$^*$} & 33.4 & 14.6 & 37.6 & 54.6 & \textbf{56.4} & \textbf{57.7} \\
  \end{tabular}

   \caption{\textbf{Image segmentation results on ADE20K \texttt{val}.}  \modelname is competitive to specialized models on ADE20K. Panoptic segmentation models use single-scale inference by default, multi-scale numbers are marked with~$^*$. For semantic segmentation, we report both single-scale (s.s.) and multi-scale (m.s.) inference results. }

\label{tab:benchmark:ade20k}
\end{table*}

\begin{table*}[t]
  \centering

  \tablestyle{5pt}{1.2}\scriptsize\begin{tabular}{l|l | x{36}x{36} | x{36}x{36}}
  & & \multicolumn{2}{c|}{panoptic model} & \multicolumn{2}{c}{semantic model} \\
  method & backbone & PQ & mIoU$_\text{pan}$ & mIoU (s.s.) & mIoU (m.s.) \\
  \shline
  \multirow{3}{*}{Panoptic-DeepLab~\cite{cheng2020panoptic}}
  & ensemble & 42.2$^*$ & 58.7$^*$ & -  & - \\
  & SWideRNet~\cite{chen2020scaling} & 43.7\phantom{$^*$} & 59.4\phantom{$^*$} & - & -  \\
  & SWideRNet~\cite{chen2020scaling} & 44.8$^*$ & 60.0$^*$ & - & -  \\
  \hline
  Panoptic FCN~\cite{li2021fully} & Swin-L$^{\text{\textdagger}}$ & \textbf{45.7}\phantom{$^*$} & -\phantom{$^*$} & - & - \\
  \hline
  MaskFormer~\cite{cheng2021maskformer} & R50 & -\phantom{$^*$} & -\phantom{$^*$} & 53.1 & 55.4 \\
  HMSANet~\cite{tao2020hierarchical} & HRNet~\cite{WangSCJDZLMTWLX19hrnet} & -\phantom{$^*$} & -\phantom{$^*$} & - & 61.1 \\
  \hline\hline
  \multirow{2}{*}{\textbf{\modelname} (ours)} &
  R50\phantom{$^{\text{\textdagger}}$} & 36.3\phantom{$^*$} & 50.7\phantom{$^*$} & 57.4 & 59.0 \\
  & Swin-L$^{\text{\textdagger}}$ & \textbf{45.5}\phantom{$^*$} & \textbf{60.8}\phantom{$^*$} & \textbf{63.2} & \textbf{64.7} \\
  \end{tabular}

   \caption{\textbf{Image segmentation results on Mapillary Vistas \texttt{val}.}  \modelname is competitive to specialized models on Mapillary Vistas. Panoptic segmentation models use single-scale inference by default, multi-scale numbers are marked with~$^*$. For semantic segmentation, we report both single-scale (s.s.) and multi-scale (m.s.) inference results. }

\label{tab:benchmark:mapillary}
\end{table*}

\section{Additional results}
\label{app:results}
Here, we provide more results of \modelname with different backbones on COCO panoptic~\cite{kirillov2017panoptic} for panoptic segmentation, COCO~\cite{lin2014coco} for instance segmentation and ADE20K~\cite{zhou2017ade20k} for semantic segmentation. More specifically, for each benckmark, we evaluate \modelname with ResNet~\cite{he2016deep} with 50 and 101 layers, as well as Swin~\cite{liu2021swin} Tiny, Small, Base and Large variants as backbones. We use ImageNet~\cite{Russakovsky2015} pre-trained checkpoints to initialize backbones.

\subsection{Panoptic segmentation.}
\label{app:results:panoptic}

In \tabref{tab:panseg:coco_full}, we report \modelname with various backbones on COCO panoptic \texttt{val2017}. \modelname outperforms \emph{all} existing panoptic segmentation models with various backbones. Our best model sets a new state-of-the-art of $57.8$ PQ.

In \tabref{tab:panseg:coco_test_dev}, we further report the best \modelname model on the \texttt{test-dev} set. Note that \modelname \textbf{trained only with the standard \texttt{train2017} data}, achieves the \emph{absolute} new state-of-the-art performance on both validation and test set. \modelname even outperforms the best COCO competition entry which uses extra training data and test-time augmentation.

\subsection{Instance segmentation.}
\label{app:results:instance}

In \tabref{tab:insseg:coco_full}, we report \modelname results obtained with various backbones on COCO \texttt{val2017}. \modelname outperforms the best single-scale model, HTC++~\cite{liu2021swin,chen2019hybrid}. Note that it is non-trivial to do multi-scale inference for instance-level segmentation tasks without introducing complex post-processing like non-maximum suppression. Thus, we only compare \modelname with other  single-scale inference models. We believe multi-scale inference can further improve \modelname performance and it remains an interesting future work.

In \tabref{tab:insseg:coco_test_dev}, we further report the best \modelname model on the \texttt{test-dev} set. \modelname achieves the \emph{absolute} new state-of-the-art performance on both validation and test set. On the one hand, \modelname is extremely good at segmenting large objects: we can even outperform the challenge winner (which uses extra training data, model ensemble, \etc) on AP$^\text{L}$ by a large margin without any bells-and-whistles. On the other hand, the poor performance on small objects leaves room for further improvement in the future.

\subsection{Semantic segmentation.}
\label{app:results:semantic}

In \tabref{tab:semseg:ade20k_full}, we report \modelname results obtained with various backbones on ADE20K \texttt{val}. \modelname outperforms \emph{all} existing semantic segmentation models with various backbones. Our best model sets a new state-of-the-art of $57.7$ mIoU.

In \tabref{tab:semseg:ade20k_test}, we further report the best \modelname model on the \texttt{test} set. Following~\cite{cheng2021maskformer}, we train \modelname on the union of ADE20K \texttt{train} and \texttt{val} set with ImageNet-22K pre-trained checkpoint and use multi-scale inference. \modelname is able to outperform previous state-of-the-art methods on all metrics.

\section{Additional datasets}
\label{app:datasets}

We study \modelname on three image segmentation tasks (panoptic, instance and semantic segmentation) using four datasets. Here we report additional results on Cityscapes~\cite{Cordts2016Cityscapes}, ADE20K~\cite{zhou2017ade20k} and Mapillary Vistas~\cite{neuhold2017mapillary} as well as more detailed training settings.

\subsection{Cityscapes}
Cityscapes is an urban egocentric street-view dataset with high-resolution images ($1024 \x 2048$ pixels). It contains 2975 images for training, 500 images for validation and 1525 images for testing with a total of 19 classes.

\noindent\textbf{Training settings.} For all three segmentation tasks: we use a crop size of $512 \x 1024$, a batch size of 16 and train all models for 90k iterations. During inference, we operate on the whole image ($1024 \x 2048$). Other implementation details largely follow Section 4.1
(panoptic and instance segmentation follow semantic segmentation training settings), except that we use 200 queries for panoptic and instance segmentation models with Swin-L backbone. All other backbones or semantic segmentation models use 100 queries.

\noindent\textbf{Results.} In \tabref{tab:benchmark:cityscapes_full}, we report \modelname results obtained with various backbones on Cityscapes for three segmentation tasks and compare it with other state-of-the-art methods \emph{without using extra data}. For panoptic segmentation, \modelname with Swin-L backbone outperforms the state-of-the-art Panoptic-DeepLab~\cite{cheng2020panoptic} with SWideRnet~\cite{chen2020scaling} using single-scale inference. For semantic segmentation, \modelname with Swin-B backbone outperforms the state-of-the-art SegFormer~\cite{xie2021segformer}.

\subsection{ADE20K}

\noindent\textbf{Training settings.} For panoptic and instance segmentation, we use the exact same training parameters as we used for semantic segmentation, except that we always use a crop size of $640 \x 640$ for all backbones. Other implementation details largely follow Section 4.1
, except that we use 200 queries for panoptic and instance segmentation models with Swin-L backbone. All other backbones or semantic segmentation models use 100 queries.

\noindent\textbf{Results.} In \tabref{tab:benchmark:ade20k}, we report the results of \modelname obtained with various backbones on ADE20K for three segmentation tasks and compare it with other state-of-the-art methods. \modelname with Swin-L backbone sets a new state-of-the-art performance on ADE20K for panoptic segmentation. As there are few papers reporting results on ADE20K, we hope this experiment could set up a useful benchmark for future research.

\subsection{Mapillary Vistas}
Mapillary Vistas is a large-scale urban street-view dataset with 18k, 2k and 5k images for training, validation and testing. It contains images with a variety of resolutions, ranging from $1024 \x 768$ to $4000 \x 6000$. We only report panoptic and semantic segmentation results for this dataset.

\noindent\textbf{Training settings.}
For both panoptic and semantic segmentation, we follow the same data augmentation of \cite{cheng2021maskformer}: standard random scale jittering between 0.5 and 2.0, random horizontal flipping, random cropping with a crop size of $1024 \x 1024$ as well as random color jittering. We train our model for 300k iterations with a batch size of 16 using the ``poly'' learning rate schedule~\cite{deeplabV2}. During inference, we resize the longer side to 2048 pixels. Our panoptic segmentation model with a Swin-L backbone uses 200 queries. All other backbones or semantic segmentation models use 100 queries.

\noindent\textbf{Results.}
In \tabref{tab:benchmark:mapillary}, we report \modelname results obtained with various backbones on Mapillary Vistas for panoptic and semantic segmentation tasks and compare it with other state-of-the-art methods. Our \modelname is very competitive compared to state-of-the art specialized models even if it is not designed for Mapillary Vistas.

\section{Additional ablation studies}
\label{app:abl}
We perform additional ablation studies of \modelname using the same settings that we used in the main paper: a single ResNet-50 backbone~\cite{he2016deep}.

\subsection{Convergence analysis}
\begin{figure*}[h!]
    \centering
    \includegraphics[width=0.4\linewidth]{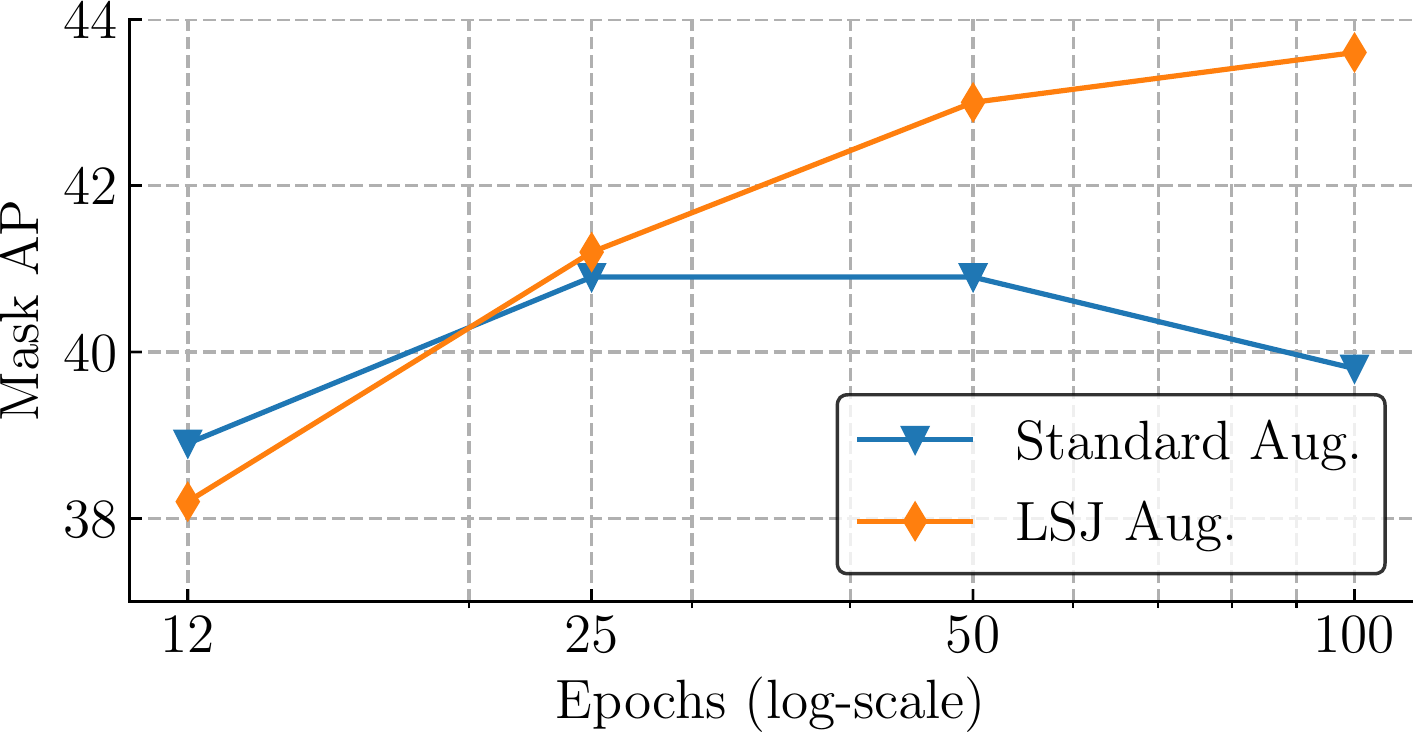}
    \caption{\textbf{Convergence analysis.} We train \modelname with different epochs using either standard scale augmentation (Standard Aug.)~\cite{wu2019detectron2} or the more recent large-scale jittering augmentation (LSJ Aug.)~\cite{ghiasi2021simple,du2021simple}. \modelname converges in 25 epochs using standard augmentation and almost converges in 50 epochs using large-scale jittering augmentation. Using LSJ also improves performance with  longer training epochs (\ie, with more than 25 epochs).}
    \label{fig:ablation:convergence}
\end{figure*}

We train \modelname  with 12, 25, 50 and 100 epochs with either standard scale augmentation (Standard Aug.)~\cite{wu2019detectron2} or the more recent large-scale jittering augmentation (LSJ Aug.)~\cite{ghiasi2021simple,du2021simple}. As shown in \figref{fig:ablation:convergence}, \modelname converges in 25 epochs using standard augmentation and almost converges in 50 epochs using large-scale jittering augmentation. This shows that \modelname with our proposed Transformer decoder converges faster than models using the standard Transformer decoder: \eg, DETR~\cite{detr} and MaskFormer~\cite{cheng2021maskformer} require 500 epochs and 300 epochs respectively.

\subsection{Masked attention analysis}

\begin{figure}[t!]
    \begin{subtable}{1.0\linewidth}
    \centering
    \includegraphics[width=\linewidth]{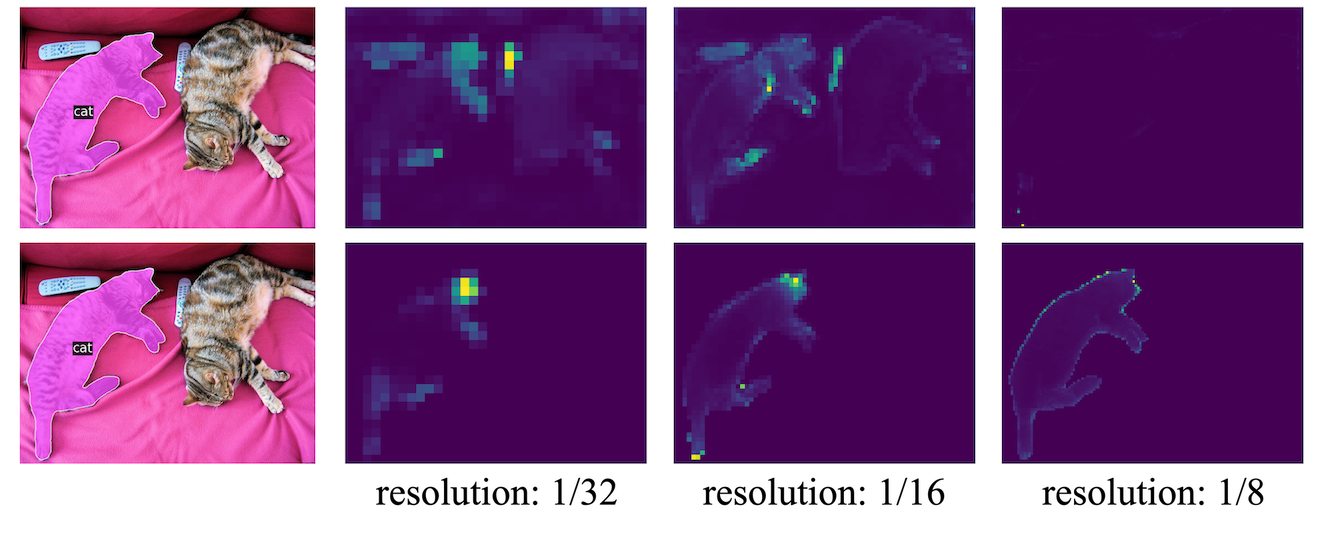}
    \vspace{-6mm}
    \caption{
        Visualization of cross-attention (top) and masked attention (bottom) for different resolutions.
    }
    \label{fig:vis_attn}
    \end{subtable}\vspace{0mm}
    \begin{subtable}{1.0\linewidth}
    \centering
    \tablestyle{2pt}{1.2}
    \scriptsize
    \begin{tabular}{l | x{18}x{18} | x{18}x{18} | x{18}x{18} | x{18}x{18} }
    & \multicolumn{2}{c|}{$1/32$} & \multicolumn{2}{c|}{$1/16$} & \multicolumn{2}{c|}{$1/8$} & \multicolumn{2}{c}{average} \\
    & fg & bg & fg & bg & fg & bg & fg & bg \\
    \shline
    cross-attention & 0.23 & 0.77 & 0.23 & 0.77 & 0.15 & 0.85 & 0.20 & 0.80 \\
    masked attention & 0.53 & 0.47 & 0.61 & 0.39 & 0.64 & 0.36 & 0.59 & 0.41 \\
    \end{tabular}

    \caption{Cumulative attention weights on foreground (fg) and background (bg) regions for different resolutions.}

    \label{tab:attn_weights_analysis}
    \end{subtable}
    \vspace{-0.3cm}
    \caption{Masked attention analysis.}
    \vspace{-0.5cm}
\end{figure}

We quantitatively and qualitatively analyzed the COCO panoptic model with the R50 backbone. First, we visualize the last three attention maps of our model using cross-attention (\figref{fig:vis_attn} top) and masked attention (\figref{fig:vis_attn} bottom) of a single query that predicts the ``cat.'' 
With cross-attention, the attention map spreads over the entire image and the region with highest response is outside the object of interest.
We believe this is because the softmax used in cross-attention never attains zero, and small attention weights on large background regions start to dominate.
Instead, masked attention limits the attention weights to focus on the object.
We validate this hypothesis in \tabref{tab:attn_weights_analysis}: we compute the cumulative attention weights on foreground (defined by the matching ground truth to each prediction) and background for all queries on the entire COCO \texttt{val} set.
On average, only $20\%$ of the attention weights in cross-attention focus on the foreground while masked attention increases this ratio to almost $60\%$.
\begin{figure}[t!]
    \centering
        \includegraphics[width=1.0\linewidth]{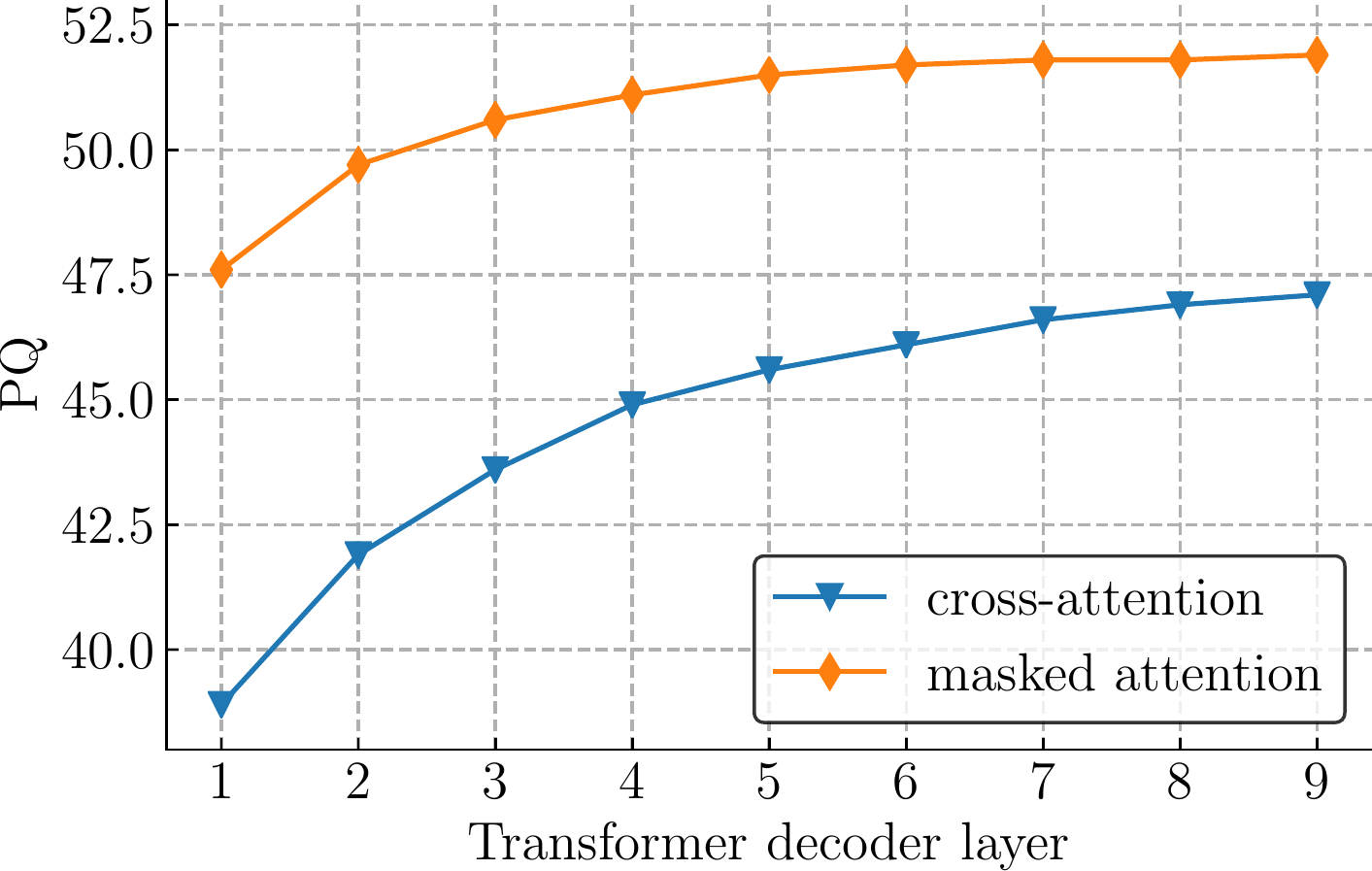}

    \caption{Panoptic segmentation performance of each Transformer decoder layer.}
    \label{fig:pq_vs_layer}
\end{figure}
Second, we plot the panoptic segmentation performance
using output from each Transformer decoder layer (\figref{fig:pq_vs_layer}). We find masked attention with a single Transformer decoder layer already outperforms cross-attention with 9 layers.
We hope the effectiveness of masked attention, together with this analysis, leads to better attention design.

\subsection{Object query analysis}
Object queries play an important role in \modelname. We ablate different design choices of object queries including the number of queries and making queries learnable.

\noindent\textbf{Number of queries.}
We study the effect of different number of queries for three image segmentation tasks in \tabref{tab:ablation:number_queries}. For instance and semantic segmentation, using 100 queries achieves the best performance, while using 200 queries can further improve panoptic segmentation results. As panoptic segmentation is a combination of instance and semantic segmentation, it has more segments per image than the other two tasks. This ablation suggests that picking the number of queries for \modelname may depend on the number of segments per image for a particular task or dataset.

\noindent\textbf{Learnable queries.}
An object query consists of two parts: object query features and object query positional embeddings. Object query features are only used as the initial input to the Transformer decoder and are updated through decoder layers; whereas query positional embeddings are added to query features in every Transformer decoder layer when computing the attention weights. In DETR~\cite{detr}, query features are zero-initialized and query positional embeddings are learnable. Furthermore, there is no direct supervision on these query features before feeding them into the Transformer (since they are zero vectors). In our \modelname, we still make query positional embeddings learnable. In addition, we make query features learnable as well and directly apply losses on these learnable query features before feeding them into the Transformer decoder.

In \tabref{tab:ablation:learnable_queries}, we compare our learnable query features with zero-initialized query features in DETR. We find it is important to directly supervise object queries even before feeding them into the Transformer decoder. Learnable queries \textit{without} supervision perform similarly well as zero-initialized queries in DETR.

\subsection{MaskFormer \vs \modelname}
\modelname builds upon the same meta architecture as MaskFormer~\cite{cheng2021maskformer} with two major differences: 1) We use more advanced training parameters summarized in \tabref{tab:ablation:maskformer_params:a}; and 2) we propose a new Transformer decoder with masked attention, instead of using the standard Transformer decoder, as well as some optimization improvements summarized in \tabref{tab:ablation:maskformer_params:b}. To better understand \modelname's improvements over MaskFormer, we perform ablation studies on training parameter improvements and Transformer decoder improvements in isolation.

In \tabref{tab:ablation:maskformer_params:c}, we study our new training parameters. We train the MaskFormer model with either its original training parameters in~\cite{cheng2021maskformer} or our new training parameters. We observe significant improvements of using our new training parameters for MaskFormer as well. This shows the new training parameters are also generally applicable to other models.

In \tabref{tab:ablation:maskformer_params:d}, we study our new Transformer decoder. We train a MaskFormer model and a \modelname model with the exact same backbone, \ie, a ResNet-50; pixel decoder, \ie, a FPN; and training parameters. That is, the only difference is in the Transformer decoder, summarized in \tabref{tab:ablation:maskformer_params:b}. We observe improvements for all three tasks, suggesting that the new Transformer decoder itself is indeed better than the standard Transformer decoder.

\begin{figure}[t!]
    \centering
        \includegraphics[width=1.0\linewidth]{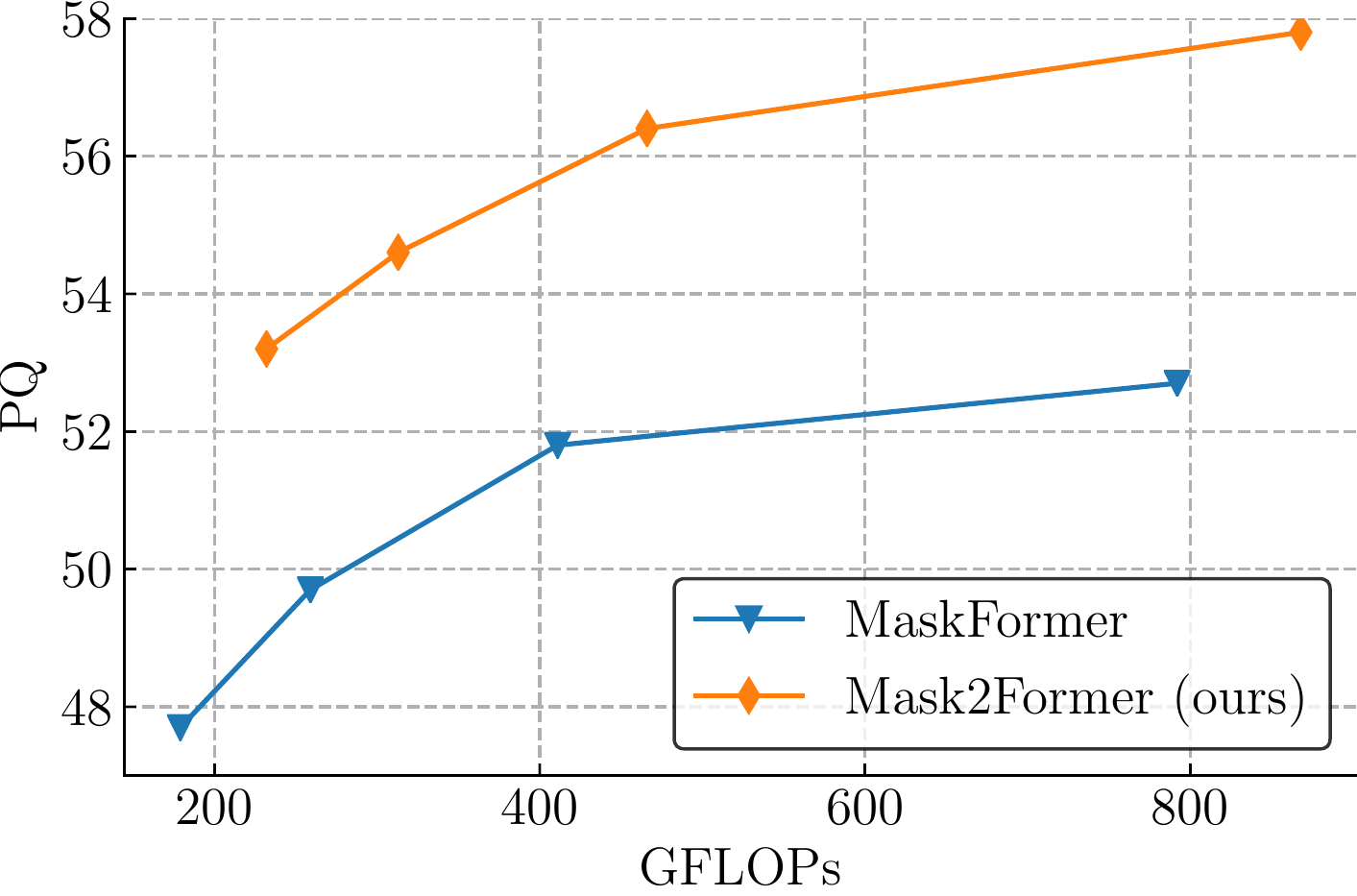}
    \caption{MaskFormer~\cite{cheng2021maskformer} \vs Mask2Former (ours) with different Swin Transformer backbones.}
    \label{fig:pq_vs_flops}
\end{figure}

While computational efficiency was not our primary goal, we find that Mask2Former actually has a better compute-performance trade-off compared to MaskFormer (\figref{fig:pq_vs_flops}). Even the lightest instantiation of Mask2Former outperforms the heaviest MaskFormer instantiation, using  $\frac{1}{4}^\text{th}$ the FLOPs.

\section{Visualization}
\label{app:vis}
\begin{figure*}[!t]
    \centering

    \begin{adjustbox}{width=\textwidth}
    \bgroup
    \def\arraystretch{0.2}
    \setlength\tabcolsep{0.2pt}
    \begin{tabular}{cccc}
    \includegraphics[height=2cm]{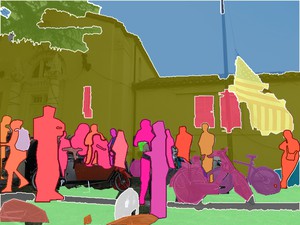} &
    \includegraphics[height=2cm]{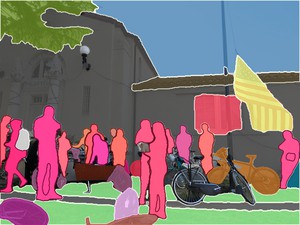} &
    \includegraphics[height=2cm]{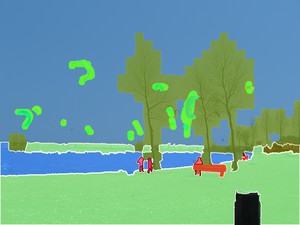} &
    \includegraphics[height=2cm]{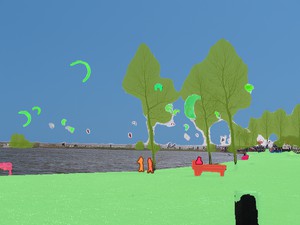} \\
    \end{tabular} \egroup
    \end{adjustbox}

    \begin{adjustbox}{width=\textwidth}
    \bgroup
    \def\arraystretch{0.2}
    \setlength\tabcolsep{0.2pt}
    \begin{tabular}{cccc}
    \includegraphics[height=2cm]{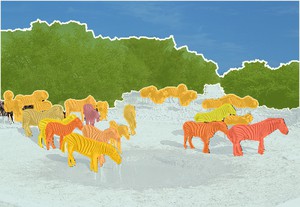} &
    \includegraphics[height=2cm]{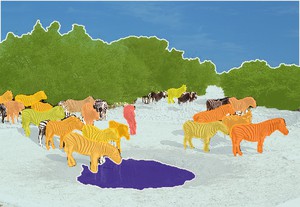} &
    \includegraphics[height=2cm]{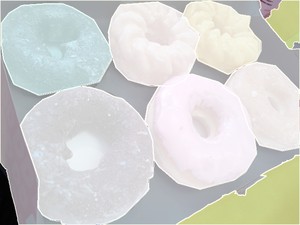} &
    \includegraphics[height=2cm]{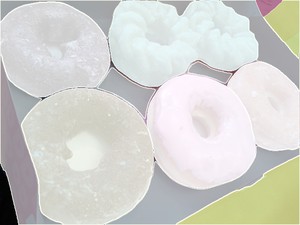} \\
    \end{tabular} \egroup
    \end{adjustbox}

    \begin{adjustbox}{width=\textwidth}
    \bgroup
    \def\arraystretch{0.2}
    \setlength\tabcolsep{0.2pt}
    \begin{tabular}{cccc}
    \includegraphics[height=2cm]{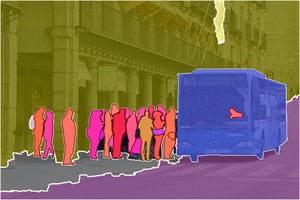} &
    \includegraphics[height=2cm]{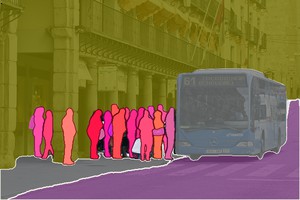} &
    \includegraphics[height=2cm]{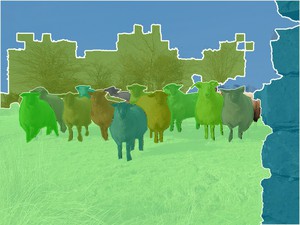} &
    \includegraphics[height=2cm]{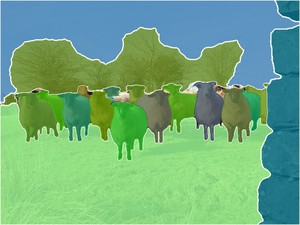} \\
    \end{tabular} \egroup
    \end{adjustbox}

    \begin{adjustbox}{width=\textwidth}
    \bgroup
    \def\arraystretch{0.2}
    \setlength\tabcolsep{0.2pt}
    \begin{tabular}{cccc}
    \includegraphics[height=2cm]{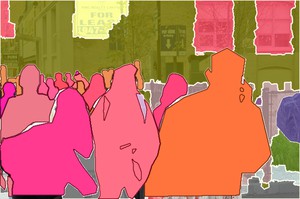} &
    \includegraphics[height=2cm]{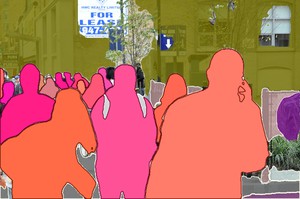} &
    \includegraphics[height=2cm]{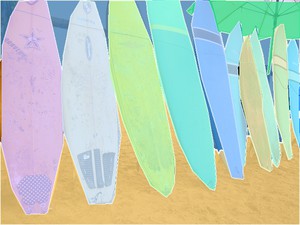} &
    \includegraphics[height=2cm]{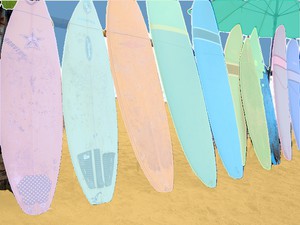} \\
    \end{tabular} \egroup
    \end{adjustbox}

    \begin{adjustbox}{width=\textwidth}
    \bgroup
    \def\arraystretch{0.2}
    \setlength\tabcolsep{0.2pt}
    \begin{tabular}{cccc}

    \includegraphics[height=2cm]{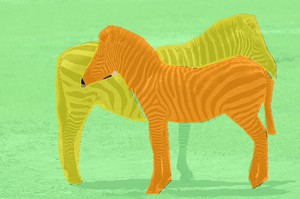} &
    \includegraphics[height=2cm]{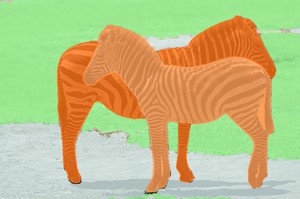} &
    \includegraphics[height=2cm]{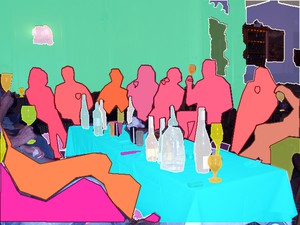} &
    \includegraphics[height=2cm]{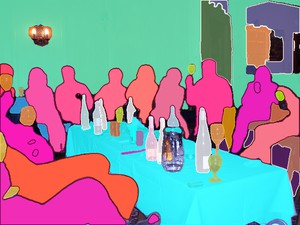} \\
    \end{tabular} \egroup
    \end{adjustbox}

    \begin{adjustbox}{width=\textwidth}
    \bgroup
    \def\arraystretch{0.2}
    \setlength\tabcolsep{0.2pt}
    \begin{tabular}{cccc}
    \includegraphics[height=2cm]{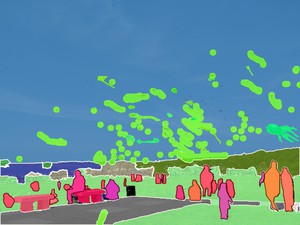} &
    \includegraphics[height=2cm]{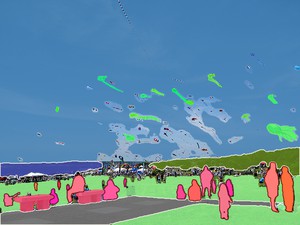} &
    \includegraphics[height=2cm]{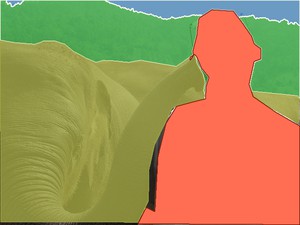} &
    \includegraphics[height=2cm]{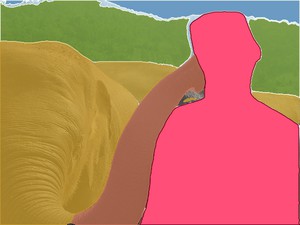} \\
    \end{tabular} \egroup
    \end{adjustbox}

  \caption{
  Visualization of \textbf{panoptic segmentation} predictions on the COCO panoptic dataset: \modelname with Swin-L backbone which achieves 57.8 PQ on the validation set. First and third columns: ground truth. Second and fourth columns: prediction. \textbf{Last row shows failure cases.}
  }
  \label{fig:vis_panseg}
\end{figure*}

\begin{figure*}[!t]
    \centering

    \begin{adjustbox}{width=\textwidth}
    \bgroup
    \def\arraystretch{0.2}
    \setlength\tabcolsep{0.2pt}
    \begin{tabular}{cccc}
    \includegraphics[height=2cm]{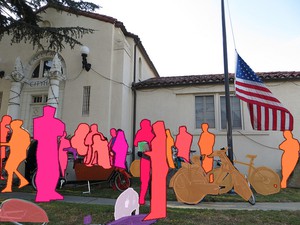} &
    \includegraphics[height=2cm]{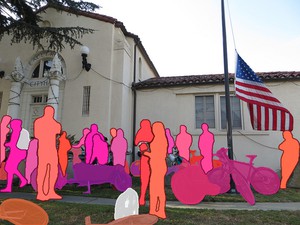} &
    \includegraphics[height=2cm]{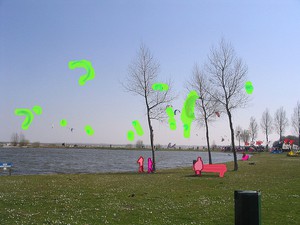} &
    \includegraphics[height=2cm]{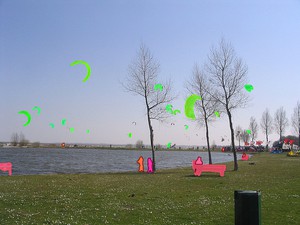} \\
    \end{tabular} \egroup
    \end{adjustbox}

    \begin{adjustbox}{width=\textwidth}
    \bgroup
    \def\arraystretch{0.2}
    \setlength\tabcolsep{0.2pt}
    \begin{tabular}{cccc}
    \includegraphics[height=2cm]{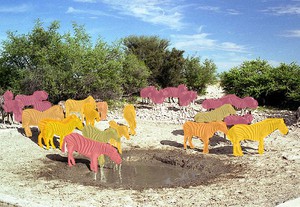} &
    \includegraphics[height=2cm]{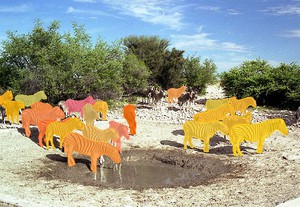} &
    \includegraphics[height=2cm]{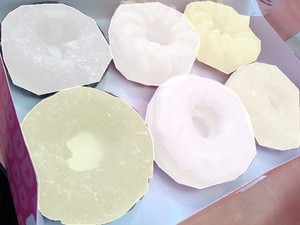} &
    \includegraphics[height=2cm]{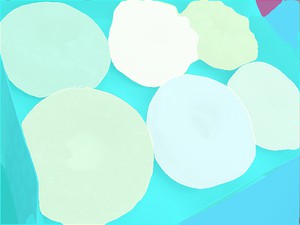} \\
    \end{tabular} \egroup
    \end{adjustbox}

    \begin{adjustbox}{width=\textwidth}
    \bgroup
    \def\arraystretch{0.2}
    \setlength\tabcolsep{0.2pt}
    \begin{tabular}{cccc}
    \includegraphics[height=2cm]{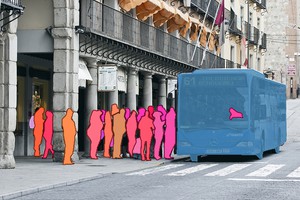} &
    \includegraphics[height=2cm]{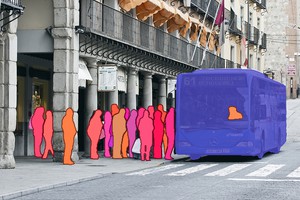} &
    \includegraphics[height=2cm]{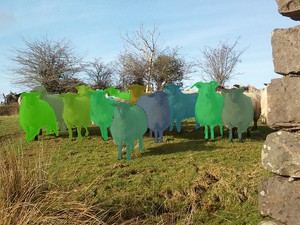} &
    \includegraphics[height=2cm]{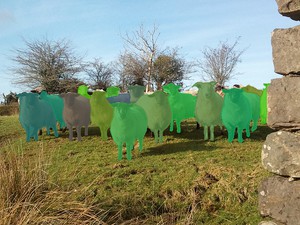} \\
    \end{tabular} \egroup
    \end{adjustbox}

    \begin{adjustbox}{width=\textwidth}
    \bgroup
    \def\arraystretch{0.2}
    \setlength\tabcolsep{0.2pt}
    \begin{tabular}{cccc}
    \includegraphics[height=2cm]{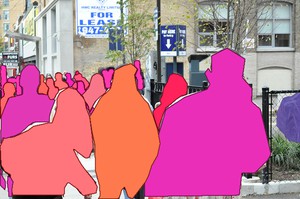} &
    \includegraphics[height=2cm]{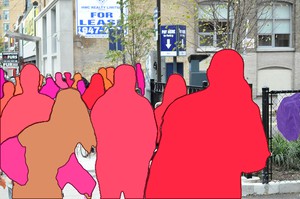} &
    \includegraphics[height=2cm]{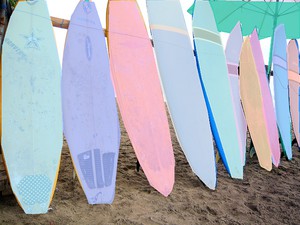} &
    \includegraphics[height=2cm]{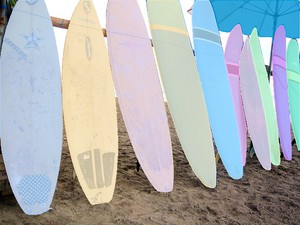} \\
    \end{tabular} \egroup
    \end{adjustbox}

    \begin{adjustbox}{width=\textwidth}
    \bgroup
    \def\arraystretch{0.2}
    \setlength\tabcolsep{0.2pt}
    \begin{tabular}{cccc}

    \includegraphics[height=2cm]{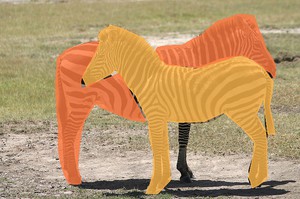} &
    \includegraphics[height=2cm]{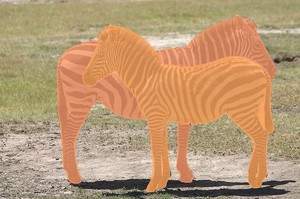} &
    \includegraphics[height=2cm]{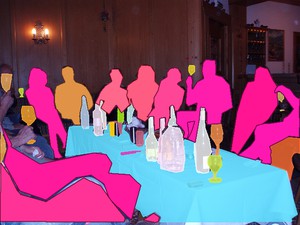} &
    \includegraphics[height=2cm]{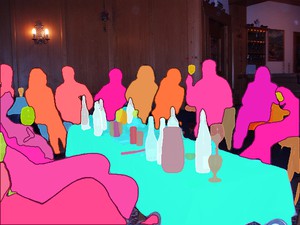} \\
    \end{tabular} \egroup
    \end{adjustbox}

    \begin{adjustbox}{width=\textwidth}
    \bgroup
    \def\arraystretch{0.2}
    \setlength\tabcolsep{0.2pt}
    \begin{tabular}{cccc}
    \includegraphics[height=2cm]{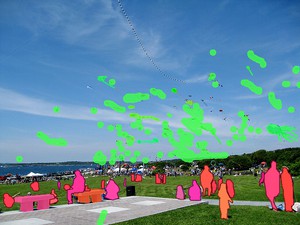} &
    \includegraphics[height=2cm]{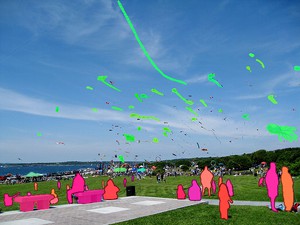} &
    \includegraphics[height=2cm]{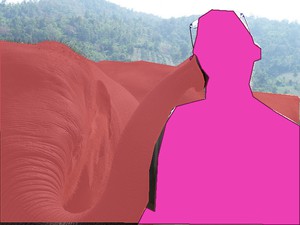} &
    \includegraphics[height=2cm]{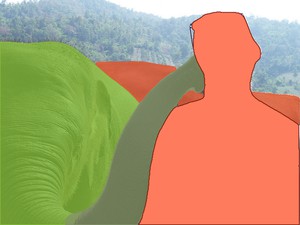} \\
    \end{tabular} \egroup
    \end{adjustbox}

  \caption{
  Visualization of \textbf{instance segmentation} predictions on the COCO dataset: \modelname with Swin-L backbone which achieves 50.1 AP on the validation set. First and third columns: ground truth. Second and fourth columns: prediction. \textbf{Last row shows failure cases.} We show predictions with confidence scores greater than 0.5.
  }
  \label{fig:vis_insseg}
\end{figure*}

\begin{figure*}[!t]
    \centering

    \begin{adjustbox}{width=\textwidth}
    \bgroup
    \def\arraystretch{0.2}
    \setlength\tabcolsep{0.2pt}
    \begin{tabular}{cccc}
    \includegraphics[height=2cm]{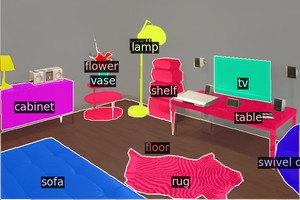} &
    \includegraphics[height=2cm]{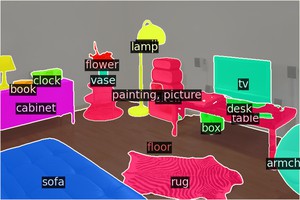} &
    \includegraphics[height=2cm]{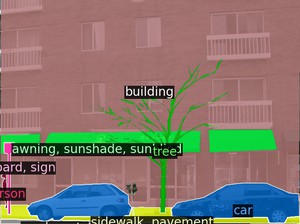} &
    \includegraphics[height=2cm]{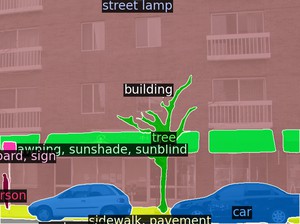} \\
    \end{tabular} \egroup
    \end{adjustbox}

    \begin{adjustbox}{width=\textwidth}
    \bgroup
    \def\arraystretch{0.2}
    \setlength\tabcolsep{0.2pt}
    \begin{tabular}{cccc}
    \includegraphics[height=2cm]{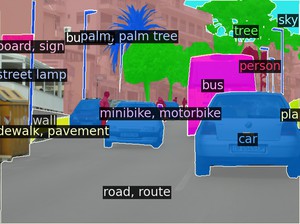} &
    \includegraphics[height=2cm]{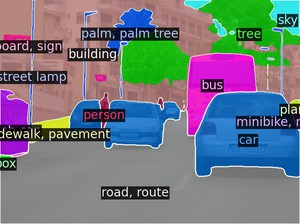} &
    \includegraphics[height=2cm]{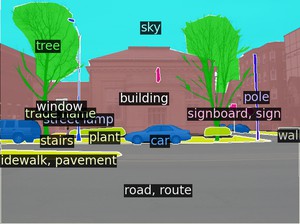} &
    \includegraphics[height=2cm]{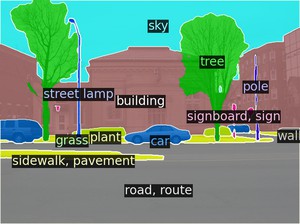} \\
    \end{tabular} \egroup
    \end{adjustbox}

    \begin{adjustbox}{width=\textwidth}
    \bgroup
    \def\arraystretch{0.2}
    \setlength\tabcolsep{0.2pt}
    \begin{tabular}{cccc}
    \includegraphics[height=2cm]{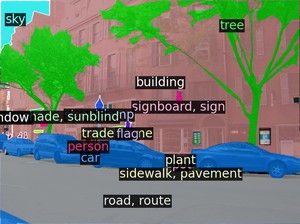} &
    \includegraphics[height=2cm]{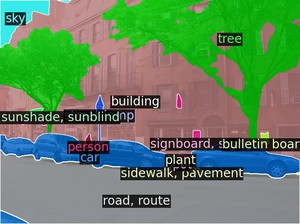} &
    \includegraphics[height=2cm]{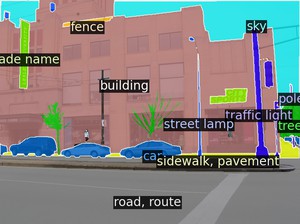} &
    \includegraphics[height=2cm]{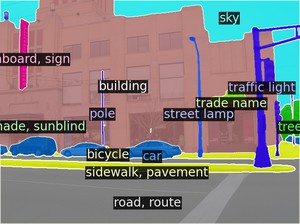} \\
    \end{tabular} \egroup
    \end{adjustbox}

    \begin{adjustbox}{width=\textwidth}
    \bgroup
    \def\arraystretch{0.2}
    \setlength\tabcolsep{0.2pt}
    \begin{tabular}{cccc}
    \includegraphics[height=2cm]{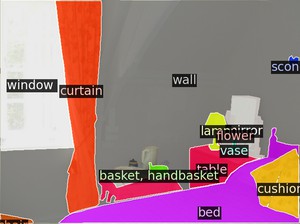} &
    \includegraphics[height=2cm]{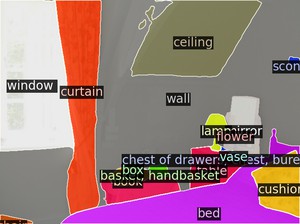} &
    \includegraphics[height=2cm]{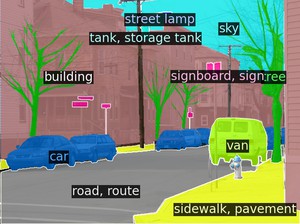} &
    \includegraphics[height=2cm]{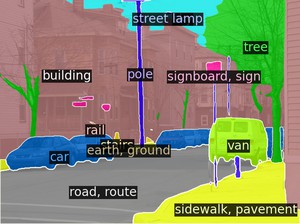} \\
    \end{tabular} \egroup
    \end{adjustbox}

    \begin{adjustbox}{width=\textwidth}
    \bgroup
    \def\arraystretch{0.2}
    \setlength\tabcolsep{0.2pt}
    \begin{tabular}{cccc}

    \includegraphics[height=2cm]{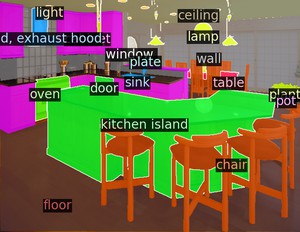} &
    \includegraphics[height=2cm]{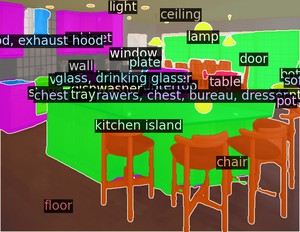} &
    \includegraphics[height=2cm]{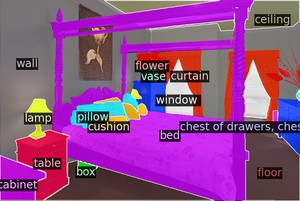} &
    \includegraphics[height=2cm]{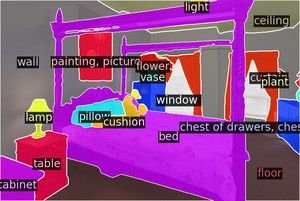} \\
    \end{tabular} \egroup
    \end{adjustbox}

    \begin{adjustbox}{width=\textwidth}
    \bgroup
    \def\arraystretch{0.2}
    \setlength\tabcolsep{0.2pt}
    \begin{tabular}{cccc}
    \includegraphics[height=2cm]{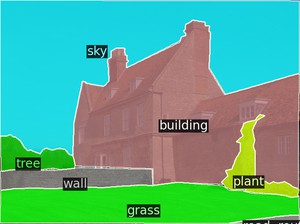} &
    \includegraphics[height=2cm]{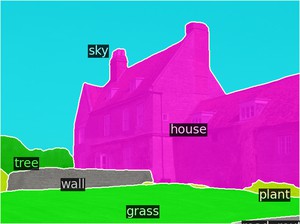} &
    \includegraphics[height=2cm]{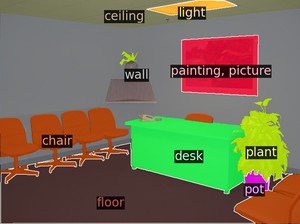} &
    \includegraphics[height=2cm]{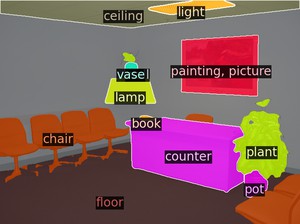} \\
    \end{tabular} \egroup
    \end{adjustbox}

  \caption{
  Visualization of \textbf{semantic segmentation} predictions on the ADE20K dataset: \modelname with Swin-L backbone which achieves 57.7 mIoU (multi-scale) on the validation set. First and third columns: ground truth. Second and fourth columns: prediction. \textbf{Last row shows failure cases.}
  }
  \label{fig:vis_semseg}
\end{figure*}

We visualize sample predictions of the \modelname model with Swin-L~\cite{liu2021swin} backbone on three tasks: COCO panoptic \texttt{val2017} set for panoptic segmentation (57.8 PQ) in \figref{fig:vis_panseg}, COCO \texttt{val2017} set for instance segmentation (50.1 AP) in \figref{fig:vis_insseg} and ADE20K validation set for semantic segmentation (57.7 mIoU, multi-scale inference) in \figref{fig:vis_semseg}.

\begin{table*}[t]
  \centering
  \begin{subtable}{0.49\linewidth}
  \centering
  \tablestyle{5pt}{1.2}
  \scriptsize
  \begin{tabular}{x{40}| x{30}x{30}x{30} | x{24}}
   & AP (COCO) & PQ (COCO) & mIoU (ADE20K) & FLOPs (COCO) \\
  \shline
  50 & 42.4 & 50.5 & 46.2 & 217G \\
  100 & \textbf{43.7} & 51.9 & \textbf{47.2} & 226G \\
  200 & 43.5 & \textbf{52.2} & 47.0 & 246G \\
  300 & 43.5 & 52.1 & 46.5 & 265G \\
  1000 & 40.3 & 50.7 & 44.8 & 405G \\
  \end{tabular}
  \caption{\textbf{Number of queries ablation.} For instance and semantic segmentation, using 100 queries achieves the best performance while using 200 queries can further improve panoptic segmentation results.
  }
  \label{tab:ablation:number_queries}
  \end{subtable}\hspace{2mm}
  \begin{subtable}{0.49\linewidth}
  \centering
  \tablestyle{5pt}{1.2}
  \scriptsize
  \begin{tabular}{l | x{30}x{30}x{30} | x{24}}
   & AP (COCO) & PQ (COCO) & mIoU (ADE20K) & FLOPs (COCO) \\
  \shline
  zero-initialized (DETR~\cite{detr}) & 42.9 & 51.2 & 45.5 & 226G \\
  learnable \textit{w/o} supervision & 42.9 & 51.2 & 47.0 & 226G \\
  learnable \textit{w/} supervision & \textbf{43.7} & \textbf{51.9} & \textbf{47.2} & 226G \\
  \multicolumn{5}{c}{~}\\
  \multicolumn{5}{c}{~}\\
  \end{tabular}
  \caption{\textbf{Learnable queries ablation.} It is important to supervise object queries before feeding them into the Transformer decoder. Learnable queries \textit{without} supervision perform similarly well as zero-initialized queries in DETR.
  }
  \label{tab:ablation:learnable_queries}
  \end{subtable}
  \caption{\textbf{Analysis of object queries.} \tabref{tab:ablation:number_queries}: ablation on number of queries. \tabref{tab:ablation:learnable_queries}: ablation on using learnable queries.
  }
  \label{tab:ablation:queries}
\end{table*}

\begin{table*}[t]
  \centering
  \begin{subtable}{1.0\linewidth}
  \centering
  \tablestyle{3pt}{1.2}
  \scriptsize
  \begin{tabular}{y{60} | x{80}x{80} }
  training parameters & MaskFormer & \modelname (ours) \\
  \shline
  learning rate & 0.0001 & 0.0001 \\
  weight decay & 0.0001 & 0.05 \\
  batch size & 16$^*$ & 16 \\
  epochs & 75$^*$ & 50 \\
  data augmentation & standard scale aug. w/ crop & LSJ aug. \\
  $\lambda_{\text{cls}}$ & 1.0 & 2.0 \\
  $\lambda_{\text{focal}}$ / $\lambda_{\text{ce}}$ & 20.0 / - & - / 5.0 \\
  $\lambda_{\text{dice}}$ & 1.0 & 5.0 \\
  mask loss & mask & 12544 sampled points \\
  \end{tabular}
  \caption{Comparison of training parameters for MaskFormer~\cite{cheng2021maskformer} and our \modelname on the COCO dataset. $^*$: in the original MaskFormer implementation, the model is trained with a batch size of 64 for 300 epochs. We find MaskFormer achieves similar performance when trained with a batch size of 16 for 75 epochs, \ie, the same number of iterations with a smaller batch size.
  }
  \label{tab:ablation:maskformer_params:a}
  \end{subtable}\vspace{2mm}
  \begin{subtable}{1.0\linewidth}
  \centering
  \tablestyle{3pt}{1.2}
  \scriptsize
  \begin{tabular}{y{60} | x{80}x{80} }
  Transformer decoder & MaskFormer & \modelname (ours) \\
  \shline
  \# of layers & 6 & 9 \\
  single layer & SA-CA-FFN & MA-SA-FFN \\
  dropout & 0.1 & 0.0 \\
  feature resolution & $\{1/32\} \x 6$ & $\{1/32, 1/16, 1/8\} \x 3$ \\
  input query features & zero init. & learnable \\
  query p.e. & learnable & learnable \\
  \end{tabular}
  \caption{Comparison of Transformer decoder in MaskFormer~\cite{cheng2021maskformer} and our \modelname. SA: self-attention, CA: cross-attention, FFN: feed-forward network, MA: masked attention, p.e.: positional embedding.
  }
  \label{tab:ablation:maskformer_params:b}
  \end{subtable}\vspace{2mm}
  \begin{subtable}{0.49\linewidth}
  \centering
  \tablestyle{4pt}{1.2}
  \scriptsize
  \begin{tabular}{y{50} y{50}| x{30}x{30}x{30}}
  \phantom{modelmodel} model & \phantom{modelmodel} training params. & AP (COCO) & PQ (COCO) & mIoU (ADE20K) \\
  \shline
  MaskFormer & MaskFormer & 34.0 \phantom{\dt{+0.0}} & 46.5 \phantom{\dt{+0.0}} & 44.5 \phantom{\dt{+0.0}} \\
  MaskFormer & \modelname & 37.8 \dt{+3.8} & 48.2 \dt{+1.7} & 45.3 \dt{+0.8} \\
  \end{tabular}
  \caption{Improvements from better \textbf{training parameters}.
  }
  \label{tab:ablation:maskformer_params:c}
  \end{subtable}
  \begin{subtable}{0.49\linewidth}
  \centering
  \tablestyle{4pt}{1.2}
  \scriptsize
  \begin{tabular}{y{60} y{40}| x{30}x{30}x{30}}
  \phantom{modelmodel} Transformer decoder & \phantom{modelmodel} pixel decoder & AP (COCO) & PQ (COCO) & mIoU (ADE20K) \\
  \shline
  MaskFormer & FPN & 37.8 \phantom{\dt{+0.0}} & 48.2 \phantom{\dt{+0.0}} & 45.3 \phantom{\dt{+0.0}} \\
  \modelname & FPN & 41.5 \dt{+3.7} & 50.7 \dt{+2.5} & 45.6 \dt{+0.3} \\
  \end{tabular}
  \caption{Improvements from better \textbf{Transformer decoder}.
  }
  \label{tab:ablation:maskformer_params:d}
  \end{subtable}
  \caption{
  \textbf{MaskFormer \vs \modelname.} \tabref{tab:ablation:maskformer_params:a} and \tabref{tab:ablation:maskformer_params:b} provide an in-depth comparison between MaskFormer and our \modelname settings. \tabref{tab:ablation:maskformer_params:c}: MaskFormer benefits from our new training parameters as well. \tabref{tab:ablation:maskformer_params:d}: Comparison between MaskFormer and our \modelname with the exact same backbone, pixel decoder and training parameters. The improvements solely come from a better Transformer decoder.
  }
  \label{tab:ablation:maskformer_params}
\end{table*}

